\documentclass[11pt,a4paper]{article}
\usepackage[margin=25mm,headheight=14pt]{geometry}
\usepackage[T1]{fontenc}
\usepackage{lmodern}
\usepackage{amsmath,amssymb}
\usepackage{longtable,booktabs,array,calc}
\usepackage{needspace}
\usepackage{graphicx}
\usepackage{xcolor}
\usepackage{hyperref}
\definecolor{linkblue}{HTML}{234E70}
\hypersetup{colorlinks=true,linkcolor=linkblue,urlcolor=linkblue,
  pdftitle={5W1H+Which: Context-Valid Semantic Indexing with Progressive Ontology Binding},
  pdfsubject={Preprint; proposed method, not empirically validated},
  pdfkeywords={5W1H, ontology binding, semantic indexing, contextual validity},
  pdfauthor={Yaxiao Liu, Pengbo Liu, Yiwen Liu, Yihua Guan, Jiaxing Song},hypertexnames=false}
\makeatletter
\renewcommand\section{\@startsection{section}{1}{\z@}%
  {-3.5ex\@plus-1ex\@minus-.2ex}{2.3ex\@plus.2ex}%
  {\normalfont\Large\bfseries\raggedright}}
\makeatother
\providecommand{\tightlist}{\setlength{\itemsep}{2pt}\setlength{\parskip}{2pt}}
\title{\bfseries 5W1H+Which: Context-Valid Semantic Indexing\\[3pt]
  \large with Progressive Ontology Binding}
\author{%
\normalsize Yaxiao Liu\textsuperscript{1}, Pengbo Liu\textsuperscript{1}, Yiwen Liu\textsuperscript{1},\\
\normalsize Yihua Guan\textsuperscript{1}, Jiaxing Song\textsuperscript{2}\\[4pt]
\small \textsuperscript{1}PwC China AI Center\quad \textsuperscript{2}Tsinghua University\\[4pt]
\small\texttt{yaxiao.y.liu@cn.pwc.com}, \texttt{rootliu@gmail.com}\\
\small\texttt{liupengbo@mails.neu.edu.cn}, \texttt{202383049@uibe.edu.cn}\\
\small\texttt{202311260036@mail.bnu.edu.cn}, \texttt{jxsong@tsinghua.edu.cn}
}
\date{\small October 24, 2026 --- v0.5 preprint revision}
\begin{document}
\maketitle
\vspace{-12pt}
\begin{quote}
\textbf{Research status.} This manuscript proposes a method, falsifiable
hypotheses, and an evaluation design. No experiments for this project
have yet been conducted. Algorithms are proposed procedures; examples
and figures illustrate mechanisms, not an implemented system, real
business facts, or measured performance. Downstream training and
evaluation are application directions, not demonstrated improvements in
agent capabilities.
\end{quote}

\section{Abstract}\label{abstract}

Semantic indexing must preserve reusable source information without
committing too early to one business schema, while providing checkable
premises for reasoning. We propose 5W1H+Which, a design that separates
source-grounded content extraction from progressive ontology binding.
The 5W1H questions organize independently addressable content units;
Which records links to versioned ontology elements, mapping relations,
applicability scopes, and validation states. Time, location, system
environment, and participant roles constrain when facts, bindings, and
rules may be used. Unbound content remains searchable, whereas formal
reasoning requires explicit premise checks. The design distinguishes
business valid time, system knowledge time, and operational traces, and
tracks dependencies for revalidation and incremental maintenance. Worked
examples illustrate cloud migration and temporal policy validity. The
resulting linked corpus can support downstream training and evaluation
organized around benchmarks, positive and negative examples,
trajectories, and costs. Trajectories preserve steps, evidence-backed
checkpoints, recovery, and local-to-global failure impact; reusable
skills remain downstream abstractions requiring cross-episode
validation. We specify falsifiable hypotheses and capability-matched
baselines to test cross-task evidence coverage, contextual misuse, and
update cost under equal information and budgets. The contribution is a
proposed indexing and maintenance contract, not a new universal ontology
or post-training algorithm. No experiments have yet been conducted, and
no performance improvement is claimed.

\textbf{Keywords:} semantic indexing; 5W1H; ontology binding; contextual
validity; knowledge updates; evidence provenance.

\section{1 Introduction}\label{introduction}

The purpose of a data index extends beyond locating a document. It
includes explaining which facts in that document can be reused, under
what conditions they hold, and which subsequent operations they can
support. A migration record may support server lookup,
maintenance-responsibility assessment, execution review, or a check of
whether a particular rule still applies. These tasks share some content,
but they need not share one business ontology or final output format.

We address a design tension: an indexer cannot enumerate all future
questions in advance, yet it must decide what information to retain. A
representation tailored too closely to the current task may hide other
uses. An overly broad description, in contrast, may omit the units,
roles, negation, versions, and scope that determine whether an answer is
valid. Ontologies clarify concepts and constraints, but mapping content
to an ontology is itself a process requiring evidence, judgment, and
maintenance; it is not inherently correct.

We model two questions separately: first, what does the source actually
express? Second, through which ontological interpretation may that
content be used in the current context? The 5W1H questions provide an
organizational entry point for the first question, while Which provides
a binding interface for the second. Here, generality means a set of
question directions shared across domains, not a fixed set of fields
sufficient for arbitrary data. Precise measurements, financial reporting
conventions, and program structure still require typed payloads and
domain extensions.

A simple example motivates this separation. A service runs on an
on-premises server this year and is scheduled to move to the cloud next
year. An initial ``on-premises operations'' label may apply within a
particular temporal and system scope. Once completion of the migration
has been verified, that label no longer applies to the new environment.
The historical record has not thereby become incorrect, nor has the
ontology's LocalServer concept disappeared. Overwriting the label
damages historical queries; failing to update it applies old rules to a
new environment; changing the actual deployment state based only on a
migration plan mistakes an intention for a fact.

The proposed contributions are threefold. First, we define content
units, Which bindings, and derived conclusions as independently
maintainable objects, preserving the queryability of unbound content.
Second, we explicitly constrain the use of bindings and rules by
context, distinguishing changes in facts, interpretations, and schemas.
Third, we specify a change-propagation mechanism and controlled
experiments to test whether this separation justifies its additional
cost. We do not claim to have invented 5W1H, ontology versioning, or
provenance, and we do not equate symbolic reasoning with absolute truth
about the world.

The longer-term objective is to support agents that adapt to new
requirements. Post-training, prompt and skill improvement, tool routing,
and workflow repair all need to connect evaluation goals, examples,
execution processes, and costs. The underlying index must therefore do
more than return isolated facts, without prematurely discarding material
around a fixed training task. It should preserve addressable content and
evidenced relations, then organize reusable material by task and goal
version. Section 9 develops this downstream use and its evaluation
direction, not a validated fourth contribution.

\section{2 Related Work and Research
Position}\label{related-work-and-research-position}

\subsection{2.1 5W1H Extraction and Event
Modeling}\label{w1h-extraction-and-event-modeling}

SocraticKG uses 5W1H-guided question answering as an intermediate
representation for text-to-knowledge-graph construction and is a direct
predecessor of our extraction component {[}R1{]}. XPEventCore
incorporates 5W1H into an event ontology and combines it with case-based
reasoning for graph enrichment, showing that the broad combination of
``5W1H plus ontology'' is not itself sufficient novelty {[}R2{]}.

The question requiring further evaluation is whether retaining
independently addressable content after extraction, deferring or
revising its business bindings, and explicitly managing applicability
contexts yields measurable benefits. This distinction is a research
position to be tested. The absence of identical terminology in prior
work does not establish that those systems cannot implement the same
capabilities.

\subsection{2.2 Dynamic Ontologies and Constrained
Mapping}\label{dynamic-ontologies-and-constrained-mapping}

OaK combines dynamic ontologies, typed functions, and task feedback,
demonstrating that ontology-based methods are not limited to manually
specified static taxonomies {[}R3{]}. Recent work on schema-ontology
mapping combines symbolic constraints with LLM subtasks to map source
schemas to ontologies, suggesting that Which should not be reduced to a
single similarity match {[}R4{]}. OM4OV examines the relationship
between ontology matching and ontology versioning, providing direct
context for deciding whether a mapping needs revalidation after its
target ontology changes {[}R5{]}.

EvoOntology further organizes schema, content, and tool layers as an
evolving ontology interface for data agents, controlling updates through
attribution-guided typed edits and paired evaluation {[}R27{]}.
Comparators should therefore include ontology revision from task
feedback, not only static industry taxonomies, and charge revision,
review, and tool-adaptation costs.

Our relationship to this work is complementary but also requires
competitive evaluation. Which can incorporate existing mapping and
version-identification methods. The additional mechanism we propose is
to make the existence of source content independent of successful
mapping, while explicitly representing the scope and lifecycle of
bindings. If a dynamic ontology system already achieves equivalent
results through the same mechanisms, the contribution must be narrowed
accordingly.

\subsection{2.3 Context, Time, and
Provenance}\label{context-time-and-provenance}

The Contextualized Knowledge Repository already treats context
organization and reasoning across contexts as explicit problems
{[}R6{]}. PROV-O provides representations for provenance relations, and
OWL-Time provides a basis for temporal entities and relations. These
standards do not automatically establish the truth of a business fact,
nor do they implement bitemporal querying or access control on behalf of
an application {[}R7, R8{]}. OWL formal semantics also requires
distinguishing ``entailed by given axioms'' from ``verified in the
world''; lack of proof cannot simply be interpreted as falsity {[}R9{]}.

We therefore do not claim to be the first to add temporal or spatial
dimensions to knowledge. The object of evaluation is whether the
combination of content retention, contextualized binding, premise
admission, and update maintenance improves index reuse and reliability
within a specified budget.

\subsection{2.4 Auditable Queries and Hypothetical
Reasoning}\label{auditable-queries-and-hypothetical-reasoning}

OntoKG-EQ uses competency questions to govern ontologies and auditable
queries, supporting the use of explicit questions to control extension
rather than adding labels without limit {[}R10{]}. WhatIfBench provides
an evaluation context for premise preservation and consistency in
counterfactual reasoning {[}R11{]}. Accordingly, we treat What-if as an
isolated scenario branch: simulated outputs must not automatically enter
the factual store, and a verbal explanation alone does not establish a
causal relationship.

These works define four direct comparison families: 5W1H extraction,
dynamic ontologies, contextual knowledge repositories, and auditable
querying. Experiments must include strong controls with these
capabilities, rather than comparing only a simple keyword index against
the complete proposed system.

\subsection{2.5 Agent Memory: Dimensions, Lifecycle, and Use-Time
Enforcement}\label{agent-memory-dimensions-lifecycle-and-use-time-enforcement}

DimMem structures memories by dimensions and routes retrieval through
constraints {[}R12{]}; dimensional indexing alone is therefore not our
novelty. RuleMem induces reusable natural-language rules {[}R13{]}, but
a retrieved rule candidate is not an approved formal premise. InMind
separates stored factual recall from implicit applicability {[}R14{]},
motivating both indirect-use tasks and negative controls where a
plausible association should not activate a memory. Recuris connects
experiential and working memory with verified harness updates {[}R15{]};
trajectory-to-skill conversion is consequently related work, not a new
contribution claimed here.

Fortunate Recall couples behavioral categories with lifecycle policies
{[}R16{]}. Its reported ablation retains generic lifecycle primitives
while removing several typed mechanisms, and a nonsignificant
correctness difference does not establish equivalence. This motivates
separating category organization from metadata and enforcement in our
own tests. \emph{Revoked but Still Authoritative} examines failures
between invalidation labels, retrieval, and action {[}R17{]}. Its
configurations require careful interpretation: the mem0 expiry
experiment uses a retrieval-flag override, whereas the default setting
withholds expired records. We do not generalize the headline to every
system's default behavior. ERRAND treats rechecking memory as budgeted
maintenance and separates in-service from abeyant records {[}R18{]}; its
controlled environment does not establish enterprise deployment results.

\Needspace{16\baselineskip}
\textbf{Table 1. Related mechanisms and the remaining test obligation.}
This is a research-position table, not an exhaustive feature audit or a
ranking of system performance. A baseline may implement our proposed
contract too.

{\begingroup\small
\begin{longtable}[]{@{}
  >{\raggedright\arraybackslash}p{(\linewidth - 4\tabcolsep) * \real{0.2400}}
  >{\raggedright\arraybackslash}p{(\linewidth - 4\tabcolsep) * \real{0.3200}}
  >{\raggedright\arraybackslash}p{(\linewidth - 4\tabcolsep) * \real{0.4400}}@{}}
\toprule\noalign{}
\begin{minipage}[b]{\linewidth}\raggedright
Prior work
\end{minipage} & \begin{minipage}[b]{\linewidth}\raggedright
Relevant mechanism
\end{minipage} & \begin{minipage}[b]{\linewidth}\raggedright
What our evaluation must isolate
\end{minipage} \\
\midrule\noalign{}
\endhead
\bottomrule\noalign{}
\endlastfoot
SocraticKG; XPEventCore {[}R1, R2{]} & 5W1H extraction and event
modeling & Organization versus extra facts or better extraction
prompts \\
OaK; OM4OV; EvoOntology {[}R3, R5, R27{]} & Dynamic ontology, mapping
versions, and feedback-driven revision & Independent content retention
and binding revalidation under equal information \\
DimMem {[}R12{]} & Dimensional memory and constrained retrieval &
Additional value beyond a dimension-aware fact graph \\
RuleMem; InMind {[}R13, R14{]} & Rule reuse and implicit applicability &
Candidate discovery versus validated admission; useful association
versus scope leakage \\
Recuris {[}R15{]} & Experience, working state, and verified updates &
Keep skill learning outside the core indexing claim \\
Fortunate Recall {[}R16{]} & Typed and generic lifecycle policies &
Category labels versus metadata, routing, and enforced policies \\
Revocation study {[}R17{]} & Invalidated records can influence retrieval
and action & Exposure, premise misuse, and unsafe action as separate
outcomes \\
ERRAND {[}R18{]} & Budgeted rechecks and temporary abeyance &
Correctness under actual total cost; no cost-free revalidation
assumption \\
JitMem; MoM {[}R28, R29{]} & Task-adaptive selection from traces;
current-state and historical memory & Read-time selection versus
write-time maintenance; cross-task view reuse and dependencies \\
MemCalib; Beyond Downstream QA {[}R30, R31{]} & Calibrated memory use;
direct evaluation of memory evolution & Stored state, memory use, and
downstream outcomes as separate measurements \\
\end{longtable}
\endgroup
}

The narrower research question is whether an explicit contract across
independently retained content, contextual bindings, admission checks,
and update dependencies improves cross-task reuse at matched cost. None
of these ingredients alone is claimed to be unprecedented. Equivalent
results from a capability-matched baseline would narrow the contribution
to an implementation or evaluation protocol.

\subsection{2.6 Stateful Testing and Compositional
Security}\label{stateful-testing-and-compositional-security}

LLM-driven multi-step security testing is already established related
work, not a novelty claim of this paper. LLM-SID combines semantic
program representations with agent-driven business-logic testing
{[}R19{]}. LogiAgent uses API relationships and multiple agents for
logical testing {[}R20{]}; its reported logical issues include both bugs
and enhancement suggestions, not exclusively security vulnerabilities.
These are particularly close comparisons for any proposed combination of
semantic indexing and testing.

MOSAIC composes knowledge-guided command behaviors against coding agents
{[}R21{]}, whereas Compositional Policy Violations studies cases where
locally compliant steps jointly violate a workflow policy {[}R22{]}. The
latter explicitly distinguishes its strict notion of composition from
stale-state and race-related failures. Our delayed-execution example
therefore motivates broader temporal policy validity; it is not
automatically an instance of that paper's narrower category.

Long-standing precedents are equally important: RESTler explores
stateful API dependencies {[}R23{]}, ChatAFL uses LLMs to guide protocol
fuzzing {[}R24{]}, PentestGPT manages context across penetration-testing
tasks {[}R25{]}, and black-box model-based work infers behavior from
traces to test application logic {[}R26{]}. Merely adding a state
machine, a timestamp, or an LLM does not distinguish our proposal.

The narrower research question is whether independently reusable
evidence, context-valid transition bindings, and counterexample-driven
revision of the indexed state abstraction improve confirmed findings and
evidence reuse under matched budgets. Neither the absence of the name
``5W1H'' nor a diagrammatic difference establishes a capability gap.
Strong implementations of these prior ideas may eliminate the proposed
advantage.

\subsection{2.7 Maintainable Corpora and Task-Adaptive
Views}\label{maintainable-corpora-and-task-adaptive-views}

JitMem retains raw interaction traces and learns to curate memory for
the current task at read time, using task feedback to improve the
curator {[}R28{]}. The same trace may yield different payloads for
different tasks. Task-conditioned evidence selection alone is therefore
not our novelty. MoM instead emphasizes a typed provenance structure
that separates current valid memory from displaced historical records
{[}R29{]}. Maintaining valid state when evidence changes and selecting
relevant evidence for a task are complementary axes, not interchangeable
solutions.

Our proposed separation of content, bindings, and view dependencies is a
testable design choice at their intersection. The question is whether it
supports reusable task views and correct repair after source or ontology
changes, relative to baselines with equivalent provenance, versioning,
selection, and maintenance capabilities. Neither richer terminology nor
a larger retained corpus establishes an advantage.

\subsection{2.8 From Memory Material to Training and Evaluation
Assets}\label{from-memory-material-to-training-and-evaluation-assets}

Agent Memory Distillation transfers teacher experience through
hierarchical external memory without updating student parameters
{[}R32{]}. Self-Retrospection Distillation instead uses retrospective
experience to construct supervision for parameter learning {[}R33{]}.
These are different meanings of distillation. An indexed corpus may
supply material to either, but improved external-memory access does not
establish a learned capability that persists without that access.

Beyond Downstream QA directly inspects memory evolution using FEB30,
while MemCalib distinguishes overuse and underuse of memory {[}R30,
R31{]}. They motivate separating stored state, evidence selection, and
use from final-answer accuracy. ImpossibleRubrics motivates an
independent correctness boundary for generated evaluation criteria
{[}R35{]}; PeakBench motivates resource-aware evaluation beyond a single
tool-call count {[}R34{]}. Together these works support studying
connected benchmark, example, trajectory, and cost views. They do not
establish that these four pillars are exhaustive or validate 5W1H+Which
itself.

\section{3 Problem Definition and Design
Boundaries}\label{problem-definition-and-design-boundaries}

\subsection{3.1 Inputs, Index, and
Queries}\label{inputs-index-and-queries}

Let \(D=\{d_i\}\) denote a set of data sources, each with a version
identifier and addressable evidence spans. Sources may be text,
structured records, or operational traces. Multimodal support requires
separate definitions of span addressing and extraction quality and is
not claimed to be solved here. The indexer produces a content set \(Z\),
a binding set \(B\), a rule set \(R\), and a derived-record set \(Y\).

A query request is represented as:

\[Q=\langle q,\tau,\kappa,c,g,a,m\rangle,\]

where \(q\) is the question; \(\tau\) is the business time being
queried; \(\kappa\) is the knowledge cutoff; \(c\) contains typed
contextual constraints such as organization, location, system
environment, and roles; \(g\) is the goal version; \(a\) is the current
access principal; and \(m\) is the response mode. Modes distinguish at
least source lookup, business-state assessment, and hypothetical
reasoning. Finding a document that asserts something does not mean that
the assertion has been verified.

Historical queries have two distinct semantics. ``What judgment was
warranted by the information known at that time?'' constrains
\(\kappa\). ``Given what is known now, what actually happened in the
past?'' uses the current knowledge state while querying a past \(\tau\).
Late-arriving evidence may revise the latter answer, but must not be
silently introduced into the former. Both must enforce current access
authorization: historical-time parameters cannot restore revoked
permissions.

\subsection{3.2 The Content Layer Is Not Six Mandatory
Strings}\label{the-content-layer-is-not-six-mandatory-strings}

5W1H is an organizational framework for extraction and browsing. A
content unit may contain only What and When, or multiple participants,
locations, and time intervals. Why or How must not be invented to fill
every field. Missing values must distinguish information not stated by
the source, a field that is inapplicable, information not yet extracted,
and information hidden by permissions. An unstated property represents
missing knowledge, not the absence of the business property itself.

A passage containing a plan, an observation, and a conjecture should be
split into separate claims rather than assigned a single confidence
score. Exact values, units, negation, conditions, quantifiers, and
modality must remain attached to their respective claims. Document-level
summaries may provide an entry point, but cannot replace claim-level
addressing.

\subsection{3.3 Probabilistic Extraction and Conditional Formal
Reasoning}\label{probabilistic-extraction-and-conditional-formal-reasoning}

The method permits an LLM to propose uncertain content and mapping
candidates, and permits checked claims to enter a rule engine. The
distinction concerns processing and evidential status, not the assertion
that ``5W1H is always probabilistic and Which is always deterministic.''
A 5W1H unit may record a verified measurement, while a Which candidate
may be incorrectly mapped.

Formal reasoning here means deriving conclusions under a declared
ontology version, rule language, and premise set. It guarantees a
derivation relationship relative to those premises and semantics, not
the completeness or truth of the inputs. An LLM's self-reported score is
only a ranking signal requiring calibration; it must not be interpreted,
without validation, as the probability that a fact is true.

\section{4 Representation: Content, Bindings, and Derived
Results}\label{representation-content-bindings-and-derived-results}

\par\medskip\noindent\begin{minipage}{\linewidth}\centering

\includegraphics[width=\linewidth]{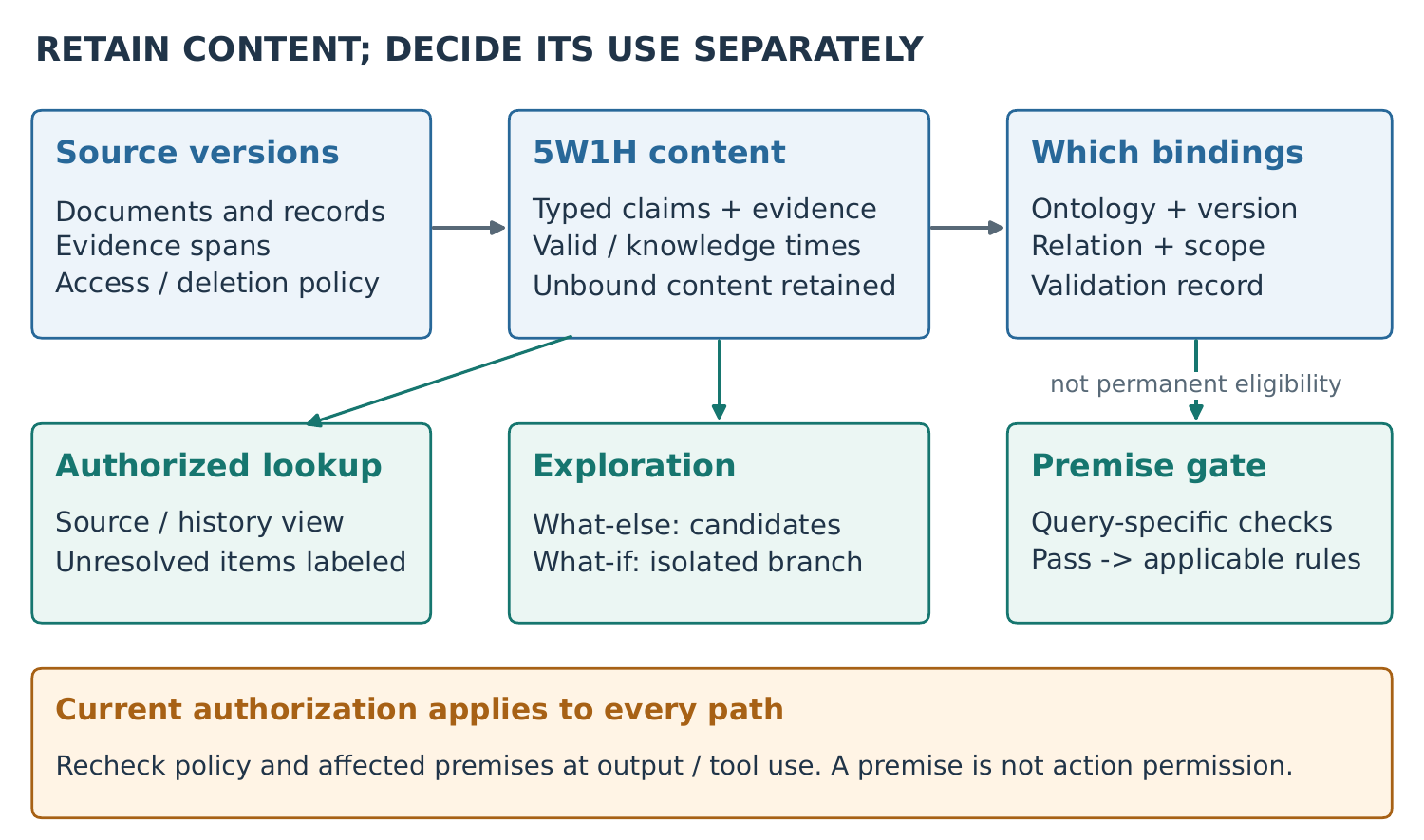}

\par\small\raggedright

\textbf{Figure 1. Proposed indexing and use-time checks.} The content
layer survives a missing Which binding. Source exploration, formal
reasoning, and hypothetical reasoning are distinct modes. Current
authorization applies to every read; output and tool actions require a
fresh policy check. The arrows describe permitted data flow, not an
implemented safety guarantee.

\end{minipage}\par\medskip

\subsection{4.1 Source-Grounded Content
Units}\label{source-grounded-content-units}

Define a content unit as:

\[z=\langle id,\chi,p,e,s,v,k\rangle,\]

where \(\chi\) contains the 5W1H organizational information; \(p\) is a
typed payload; \(e\) contains evidence references; \(s\) records
epistemic status and validation; \(v\) denotes business-valid scope; and
\(k\) is the record interval during which this version is visible in the
system. Epistemic status distinguishes source assertions, observations,
plans, inferences, and hypotheses. Validation status is a separate
field, not a substitute for epistemic status. A source's assertion that
something ``has been completed'' still requires verification appropriate
to the task.

\Needspace{16\baselineskip}
\textbf{Table 2. Content directions and distinctions that must survive
extraction.} Unknown fields remain unknown; typed payloads extend these
directions.

{\begingroup\small
\begin{longtable}[]{@{}
  >{\raggedright\arraybackslash}p{(\linewidth - 4\tabcolsep) * \real{0.1200}}
  >{\raggedright\arraybackslash}p{(\linewidth - 4\tabcolsep) * \real{0.4200}}
  >{\raggedright\arraybackslash}p{(\linewidth - 4\tabcolsep) * \real{0.4600}}@{}}
\toprule\noalign{}
\begin{minipage}[b]{\linewidth}\raggedright
Direction
\end{minipage} & \begin{minipage}[b]{\linewidth}\raggedright
Content-layer representation
\end{minipage} & \begin{minipage}[b]{\linewidth}\raggedright
Meanings that must remain distinct
\end{minipage} \\
\midrule\noalign{}
\endhead
\bottomrule\noalign{}
\endlastfoot
What & Objects, events, relations, or claims with exact payloads & A
document topic is not a specific claim that can be assessed as true or
false. \\
Who & References to participants with explicit roles & Author, operator,
responsible party, approver, and current user are different roles. \\
When & Instants, intervals, precision, time zone, and the basis for
resolving relative times & Event occurrence, planned validity,
ingestion, and publication are different times. \\
Where & Typed physical, organizational, and computational scopes & A
server room, department, cluster, and location within a source are not
one kind of place. \\
Why & Source-stated purposes and reasons, or separately recorded causal
hypotheses & A stated reason is not a demonstrated causal mechanism. \\
How & Methods, tools, procedures, parameters, and execution references &
A planned method is not necessarily the method actually executed. \\
\end{longtable}
\endgroup
}

Evidence \(e\) includes at least a source identifier, a version or
digest hash, a span location, and an access-policy reference. A hash
helps identify a version but cannot recover deleted source material.
Authorized snapshots and evidence-availability status should be managed
separately. Structured inputs may preserve rows, columns, keys, and
schema versions without first being rewritten as natural language and
re-extracted. Index entries, summaries, and embeddings of sensitive
content are subject to access restrictions as well.

\subsection{4.2 Which: Explicit Ontology
Binding}\label{which-explicit-ontology-binding}

Define a binding as:

\[b=\langle bid,zid,o^{(r)},t,\mu,\sigma,e_b,j_b\rangle,\]

where \(o^{(r)}\) is a specific ontology version; \(t\) identifies an
element in that version; \(\mu\) is the mapping type; \(\sigma\) is the
applicability context; \(e_b\) is the binding evidence; and \(j_b\) is
the validation record. A binding also maintains its own system version
and valid scope; the content unit's time information alone is
insufficient. Mappings may be one-to-many or many-to-many, but ambiguity
and competing interpretations must be recorded explicitly.

Relations such as \texttt{instanceOf}, \texttt{equivalentTo},
\texttt{narrowerThan}, and \texttt{relatedTo} are not interchangeable. A
source instance must not be equated with an ontology class merely
because their names are similar. \texttt{relatedTo} can support
exploratory navigation, but does not automatically authorize
substitution in reasoning as if equivalence had been established. The
available semantics of mapping types are defined by the adapter and rule
language, not by their string names.

Binding states may include candidate, validated, requiring revalidation,
rejected, and retired. If no suitable binding exists, the unit retains
an empty binding rather than losing its content or being forced into the
nearest category. Failure reasons distinguish at least insufficient
evidence, unresolved ambiguity, an ontology irrelevant to the current
content, mapper error, and a possible concept gap. Only the last of
these constitutes a candidate for ontology extension.

Which refers specifically to ontology binding in this manuscript. Source
references, contextual associations, process links, and conclusion
dependencies are stored separately rather than grouping all relations
under Which. This separation makes it possible to test the effect of the
binding mechanism itself.

\subsection{4.3 Context Space and Three Levels of
Change}\label{context-space-and-three-levels-of-change}

Context may be written as
\(c=\langle\tau,location,organization,environment,roles,g\rangle\).
These are heterogeneous, potentially missing, hierarchical, typed
coordinates; they need not be embedded in a numerical space with a
single distance function. ``Higher-dimensional'' means a richer set of
checkable constraints here, not an automatic improvement in vector
retrieval.

Three types of change must be maintained separately:

\begin{enumerate}
\def\labelenumi{\arabic{enumi}.}
\tightlist
\item
  \textbf{World-state change:} a service moves from on-premises to cloud
  deployment, changing the valid scope of instance facts.
\item
  \textbf{Interpretation change:} an old binding of the same record is
  found to be inaccurate, or a different binding applies under a new
  business scope.
\item
  \textbf{Ontology-schema change:} classes, properties, constraints, or
  identifiers change between versions, potentially requiring binding
  migration or revalidation.
\end{enumerate}

Rule applicability may also change independently. For example,
reassignment of maintenance responsibility does not necessarily change
deployment type. Different changes may coincide, but their distinctions
must not be collapsed into a single ``label update.'' In particular,
retirement of one on-premises server does not imply that the LocalServer
class must be deleted from the new ontology.

\subsection{4.4 Derived Records and Support
Sets}\label{derived-records-and-support-sets}

A derived result is represented as
\(y=\langle claim,\mathcal{P},c_y,status\rangle\). Here, \(\mathcal{P}\)
is a set of support records, each referencing the content versions,
binding versions, rule versions, and applicability conditions used. The
same conclusion may have multiple independent support paths. Failure of
one path does not necessarily make the entire conclusion false.

We use statuses such as ``not currently admissible for continued use,''
``supported by another valid path,'' and ``requires recomputation,''
rather than treating withdrawal of evidence as proof of the opposite
proposition. Conflicting inputs enter an explicit conflict set. The
initial prototype is not intended to feed contradictory premises into a
classical reasoner without checks, which could otherwise allow arbitrary
conclusions from inconsistency.

\subsection{4.5 Optional Transition and Trajectory
Records}\label{optional-transition-and-trajectory-records}

For authorized workflow testing, we extend rather than replace the
content model. A transition record links a precondition hypothesis, an
action, an observed response or effect, a postcondition hypothesis, and
their evidence units. Its context includes principal and role, tenant
and resource identity, service boundary, software and policy versions,
and applicable temporal intervals. A trajectory links these records
through observed ordering and explicit happens-before relations.
Wall-clock proximity alone does not establish causality or a total order
in a distributed system.

The observed response is not the hidden system state. For example, an
acknowledgment establishes receipt, not completion, authorization at
execution, or a durable state change. Competing state abstractions
remain versioned hypotheses; an authorized counterexample can require
splitting or merging states and revalidating affected bindings. This is
an operational-model proposal, not a claim to recover a complete or
minimal finite-state machine from arbitrary logs.

Which can bind transition components to a versioned workflow or
authorization ontology. However, an inferred behavioral model describes
what the system did, whereas a separately approved policy specifies what
it should permit. Learning both the behavior and its normative oracle
from the same unreviewed trace risks normalizing an existing defect.
Policy provenance, approval, enforcement boundary, and any allowed
propagation delay must therefore be separately recorded.

\section{5 Method: Progressive Binding and Context-Constrained
Queries}\label{method-progressive-binding-and-context-constrained-queries}

\subsection{5.1 Content Extraction and
Retention}\label{content-extraction-and-retention}

Indexing first identifies source versions and spans, splits compound
narratives into addressable claims, and then extracts 5W1H
organizational information and typed payloads. An LLM proposes
candidates; independent checks verify formats, units, time resolution,
entity identifiers, and citation locations. Asking the same model again
whether an answer is correct is not independent factual verification.
High-risk claims additionally require authorized external records or
human review.

The extractor must preserve ``unknown.'' A measurement record with no
stated purpose may have no Why; an operational record describing
execution steps may not establish their eventual outcome. Field coverage
is not an objective on its own, since optimizing it can encourage the
model to fill the index with guesses. Content supported by a source but
not accommodated by the current business ontology retains its
identifier, evidence, and reason for remaining unbound.

Retention does not mean extracting everything without a budget, nor does
it guarantee zero information loss. An implementation must declare its
span-selection policy, unit limits, and storage and computation budgets,
and retain original sources that can be located again. Failure analysis
must distinguish ``not extracted,'' ``extracted incorrectly,''
``discarded during binding,'' and ``not retrieved by the query.''

\subsection{5.2 Which Candidate Generation and
Validation}\label{which-candidate-generation-and-validation}

The system generates ontology candidates from claim content and context,
then checks object identity, the instance/type distinction, relation
direction, units, negation and modality, target version, and
applicability scope. Name similarity is only a candidate-generation
signal, not evidence of semantic equivalence. Different ontologies may
provide different interpretations, but incompatible interpretations must
not serve as unconditional premises within the same reasoning step.

A binding result includes its target identifier, mapping type, evidence,
checks, failure reasons, and revalidation triggers. A candidate-ranking
score and ``conditions for use satisfied'' are distinct fields. A
binding may support a specified rule only after passing the checks
required by the current task. Other candidates may remain visible in an
exploration interface, with uncertainty clearly labeled. Domain-adapter
and human-confirmation costs are included in total cost rather than
treated as free inputs.

\subsection{5.3 Premise Admission and Formal
Reasoning}\label{premise-admission-and-formal-reasoning}

For a combination of content, binding, and rule, define admission as:

\[\operatorname{Eligible}(z,b,r\mid Q)=E\land K\land V\land C\land B\land R\land A.\]

Here, \(E\) checks evidence availability and verification requirements;
\(K\) checks the knowledge cutoff; \(V\) checks business-valid scope;
\(C\) checks contextual compatibility; \(B\) checks the binding; \(R\)
checks rule applicability; and \(A\) checks current authorization. Each
check returns pass, fail, or unknown. An unknown result on a required
check excludes the item from a reasoning path requiring established
premises. Authorized users may nevertheless retrieve the source and
unresolved candidates. Being unavailable for formal reasoning must not
be confused with the nonexistence of data.

A source assertion may support ``this document asserts X'' without
sufficiently supporting business fact X. The two should compile to
distinct predicates. Joins across multiple units must also check their
shared context: two individually valid units need not refer to the same
device, measurement convention, or time interval. Temporal persistence,
identity merging, and unit conversion must not be assumed implicitly.

Let the admitted premises compile to \(K_Q\) and applicable rules form
\(R_Q\). Formal results then satisfy:

\[K_Q\cup R_Q\models_{\mathcal L} y.\]

The initial prototype is planned to use a positive, function-free
Datalog subset over a finite domain of constants, making closure
computation and update comparisons checkable. OWL ontologies from prior
work may provide types and relations, but require explicit adaptation.
We do not claim support for arbitrary OWL, conflicting axioms,
nonmonotonic rules, or rules generating unbounded new objects. Explicit
negation may be represented as a separate claim; ``not found'' does not
automatically compile to logical negation. Applications requiring a
closed-world assumption must separately state its scope and the evidence
for completeness.

Before considering exploratory results, we distinguish three decisions.
A record may be \textbf{retrievable} in an authorized historical or
unresolved-content view without being \textbf{admissible as a current
premise}. Even an admissible premise does not itself \textbf{authorize
an external action}. Semantic invalidation and access revocation are
different: the former can leave a historical record readable, while the
latter can prohibit even that read. A validated binding is also not a
permanent eligibility flag; eligibility is recomputed for the query's
time, scope, rule version, and current principal. Retrieval, generation,
and tool execution may be separated in time, so the output/tool boundary
must recheck current policy and affected premise versions. A changed
dependency requires revalidation or abstention, not merely a warning
inside a generated answer.

\subsection{5.4 What-else: Finding Omitted Content and
Interpretations}\label{what-else-finding-omitted-content-and-interpretations}

A What-else query proposes additional content from units unused by the
current task, unbound units, and neighboring evidence. Its outputs fall
into three categories: omitted claims directly supported by existing
evidence, new relation candidates requiring verification, and concept
candidates that may require ontology extension. These must not be merged
into an undifferentiated list of ``new knowledge.''

For example, descriptions of cooling, energy consumption, or maintenance
windows in a migration record may be excluded from the current
deployment ontology but remain useful for later tasks. The system should
be able to revisit them when new questions arise. However, finding an
unbound term is not equivalent to discovering a new concept. Synonyms,
extraction errors, and existing ontology elements missed during
retrieval must first be ruled out. Accepting a candidate creates a new
version and justification rather than overwriting earlier judgments.

\subsection{5.5 What-if: Isolated Hypothetical
Branches}\label{what-if-isolated-hypothetical-branches}

A hypothetical branch is denoted by
\(F=\langle base,\Delta,assumptions,scope,outputs\rangle\). It
references a fixed base snapshot, replaces premises only within a
declared scope, and preserves the remaining premises. Outputs must
identify which conclusions follow from existing facts, which depend on
added assumptions, and which conditions remain unsatisfied.

``If the service moves to the cloud next year, how will the maintenance
process change?'' can be explored using a hypothetical deployment state
and explicit responsibility rules. Such reasoning neither proves that
migration occurred nor automatically establishes that the cloud provider
assumes all responsibility. Only subsequent independent evidence can
establish a factual record, with provenance links explaining its
relationship to the earlier hypothesis. Rule-based conditional
derivation is also distinct from identifying causal effects, which
requires a separately specified causal model, intervention semantics,
and identification assumptions.

\subsection{5.6 Change Propagation and Conservative
Fallback}\label{change-propagation-and-conservative-fallback}

A change may remove support for an old conclusion or introduce a
conclusion with no previous support path. Merely following existing
dependencies to mark downstream results invalid is therefore
insufficient. The system also needs forward derivation triggered by
added predicates and rule entry points. We propose the following
maintenance procedure.

\begin{enumerate}
\def\labelenumi{\arabic{enumi}.}
\tightlist
\item
  \textbf{Version the change.} Given a consistent snapshot \(S\) and a
  change \(\Delta\) to content, bindings, ontologies, rules, or
  authorization, classify the change and create new versions. Retain
  historical records when permitted; do not overwrite their semantics in
  place.
\item
  \textbf{Revalidate bindings.} Identify affected bindings and checks,
  marking bindings as requiring revalidation when necessary.
\item
  \textbf{Reassess support.} Follow evidence and rule dependencies to
  mark affected support paths and check alternative support.
\item
  \textbf{Discover new conclusions.} Trigger derivation from added or
  revised predicates and rule entry points.
\item
  \textbf{Handle recursion.} Rederive recursive dependency components
  from grounded support to prevent circular self-support.
\item
  \textbf{Fall back conservatively.} If dependencies are incomplete,
  semantics exceed the supported fragment, or the affected scope cannot
  be reliably bounded, recompute the full relevant context slice. If
  even that slice cannot be bounded, recompute the finite knowledge
  base.
\item
  \textbf{Publish atomically.} Publish a new snapshot atomically and
  record invalidated, retained, added, and pending-revalidation results.
\end{enumerate}

Permitted incremental optimizations must agree with full recomputation
on the same new snapshot. This target depends on consistent versioned
snapshots, complete dependency and trigger indexes, deterministic
context selection, and supported finite rule semantics. This manuscript
does not prove correctness for a general language. The initial
implementation should first establish a correct full-recomputation path
and differential tests, then optimize incremental maintenance.

Authorization revocation also requires immediately blocking unauthorized
content in caches, summaries, and derived results, rather than waiting
for background maintenance. If evidence is lawfully deleted, record its
unavailability or the minimum audit metadata that may be retained.
Provenance requirements do not justify indefinite retention of sensitive
source material.

\par\medskip\noindent\begin{minipage}{\linewidth}\centering

\includegraphics[width=\linewidth]{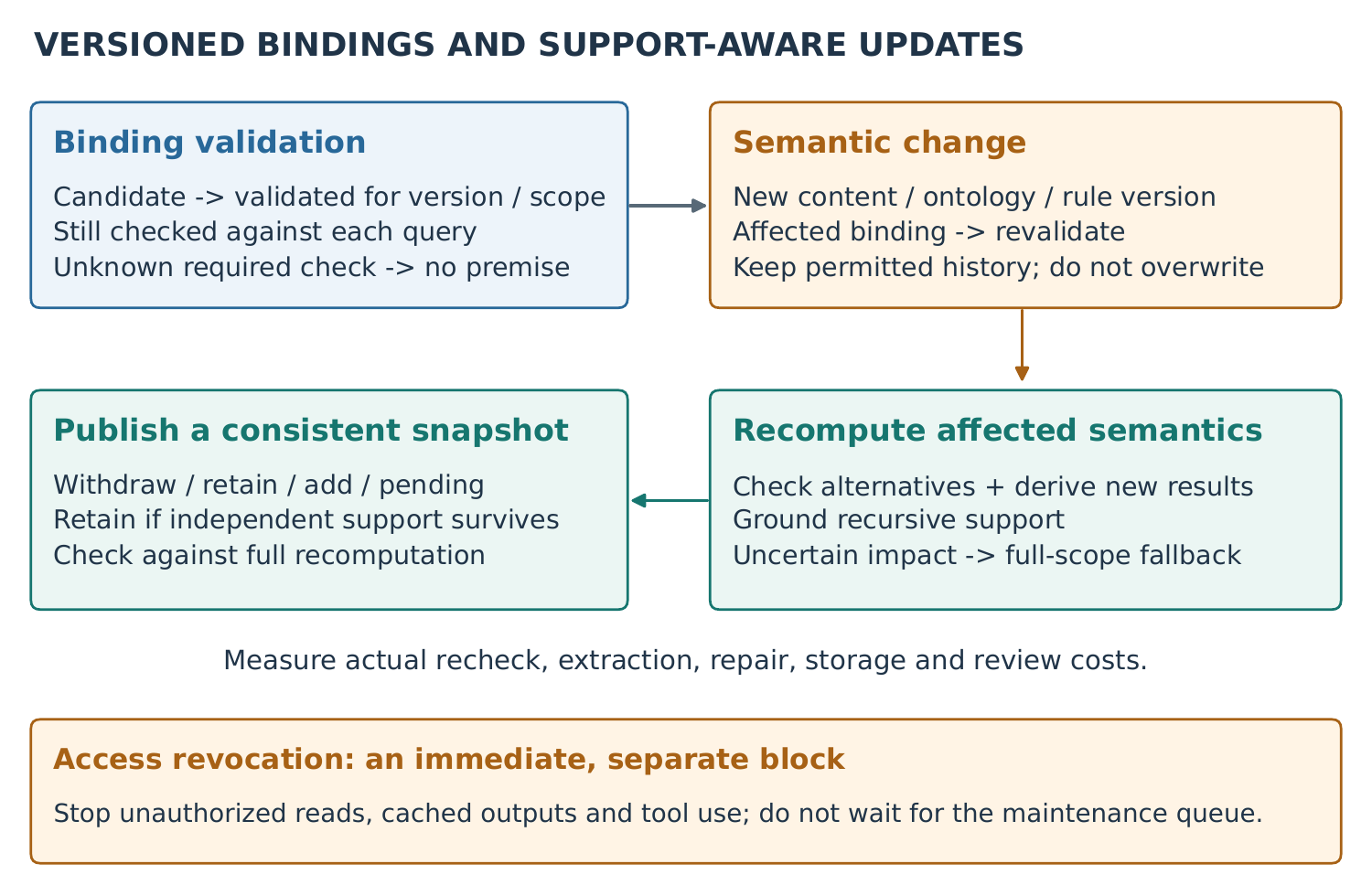}

\par\small\raggedright

\textbf{Figure 2. Binding lifecycle and update contract.} Validation is
relative to a version and scope, not a timeless truth label. Semantic
changes trigger support-aware recomputation; an independent support can
preserve a conclusion. Access revocation uses a separate immediate
block, never a delayed maintenance queue. Historical retention remains
subject to authorization and deletion policy.

\end{minipage}\par\medskip

\subsection{5.7 Maintenance Budget and Temporary
Non-Use}\label{maintenance-budget-and-temporary-non-use}

An unresolved change can put a binding into \texttt{revalidate} without
deleting its content. Required unknown checks then block the affected
formal path; authorized exploration can still report the uncertainty.
This is neither proof of falsity nor permission to continue serving the
old premise. Inspired by the separation of maintenance and use in
{[}R18{]}, we charge extraction, mapping, source rechecks, dependency
repair, storage, and human review to the maintenance ledger. Actual
expenditure, not only a shared nominal budget cap, must be reported.
Permission revocation cannot wait for a favorable maintenance score.

Tool receipts may supply new evidence about a named entity, operation,
predicate, and time. They do not verify every claim in the surrounding
conversation, and extracting or validating them has a cost even when a
tool call was already required by the task. The first implementation
tests full versus incremental maintenance; learned recheck scheduling is
a later extension, not another claimed contribution of this draft.

\section{6 Constructed Examples: Changing Deployments and Temporal
Policies}\label{constructed-examples-changing-deployments-and-temporal-policies}

\subsection{6.1 From On-Premises to Cloud
Deployment}\label{from-on-premises-to-cloud-deployment}

All dates, systems, and rules below are constructed examples, not
descriptions of a real business. The example concerns only the
production environment of service S; development environments and other
services do not change with it.

\Needspace{16\baselineskip}
\textbf{Table 3. Constructed migration evidence.} Plans, observations,
completion records, and responsibility rules have different evidential
roles.

{\begingroup\small
\begin{longtable}[]{@{}
  >{\raggedright\arraybackslash}p{(\linewidth - 6\tabcolsep) * \real{0.0700}}
  >{\raggedright\arraybackslash}p{(\linewidth - 6\tabcolsep) * \real{0.2300}}
  >{\raggedright\arraybackslash}p{(\linewidth - 6\tabcolsep) * \real{0.3500}}
  >{\raggedright\arraybackslash}p{(\linewidth - 6\tabcolsep) * \real{0.3500}}@{}}
\toprule\noalign{}
\begin{minipage}[b]{\linewidth}\raggedright
Input
\end{minipage} & \begin{minipage}[b]{\linewidth}\raggedright
Source and system-record time
\end{minipage} & \begin{minipage}[b]{\linewidth}\raggedright
Business meaning
\end{minipage} & \begin{minipage}[b]{\linewidth}\raggedright
Index treatment
\end{minipage} \\
\midrule\noalign{}
\endhead
\bottomrule\noalign{}
\endlastfoot
A & Verification record, 2026-09-01 & S runs on on-premises device L on
that date. & Record the evidenced state for that date; continued
validity requires a separately declared persistence policy. \\
B & Planning record, 2026-09-10 & S is scheduled to migrate to the cloud
on 2027-01-01. & Mark as a plan; do not change actual deployment facts
in advance. \\
C & Completion record ingested only on 2027-01-05 & Migration is
verified as effective on 2027-01-03; the former production instance is
retired. & Preserve both actual effective time and late-arriving
knowledge time. \\
D & Responsibility rule verified separately from C & After migration,
team T manages the application layer; infrastructure responsibility
follows the contract. & Do not infer all responsibilities automatically
from the CloudDeployment type. \\
\end{longtable}
\endgroup
}

For a query about deployment on 2026-09-01, A can support on-premises
deployment within the scope of its evidence. For ``Was migration
complete as of 2027-01-02?'', B still establishes only that a plan
existed. Without new evidence or a declared state-persistence
assumption, the system should return unverified rather than treat the
planned date as the completion date.

A query about the state on 2027-01-04 likewise has two interpretations.
With a knowledge cutoff of 2027-01-04, C is unavailable and cannot
establish actual completion. With knowledge acquired after 2027-01-05,
looking back at 2027-01-04, C can support that cloud deployment was
already effective. This distinction arises from knowledge time, not the
number of date fields in the source content.

Which can admit A's claim to relevant queries through an on-premises
interpretation in a Deployment ontology, and C's claim to later queries
through a cloud-deployment interpretation. Old bindings continue to
support historical questions rather than being erased. Suppose ontology
v2 subsequently subdivides cloud deployment into managed and
self-managed categories. If existing evidence does not establish the
subtype, the binding should remain at the broader type or require
revalidation; an ontology upgrade cannot automatically create more
detailed facts.

Manufacturing provides a second intended test setting. The same
measurement may support different conformity judgments under different
workstations, product versions, and tolerance rules. In a constructed
test where a tolerance changes from 0.20 mm to 0.10 mm, the raw
measurement remains unchanged; rule applicability and the derived
judgment change. These numbers are test values, not industry standards.
Finance may introduce extensions for currency, period, entity scope, and
reporting conventions, but their definitions require domain
verification. This manuscript does not elevate them directly into
accounting or compliance conclusions.

\par\medskip\noindent\begin{minipage}{\linewidth}\centering

\includegraphics[width=\linewidth]{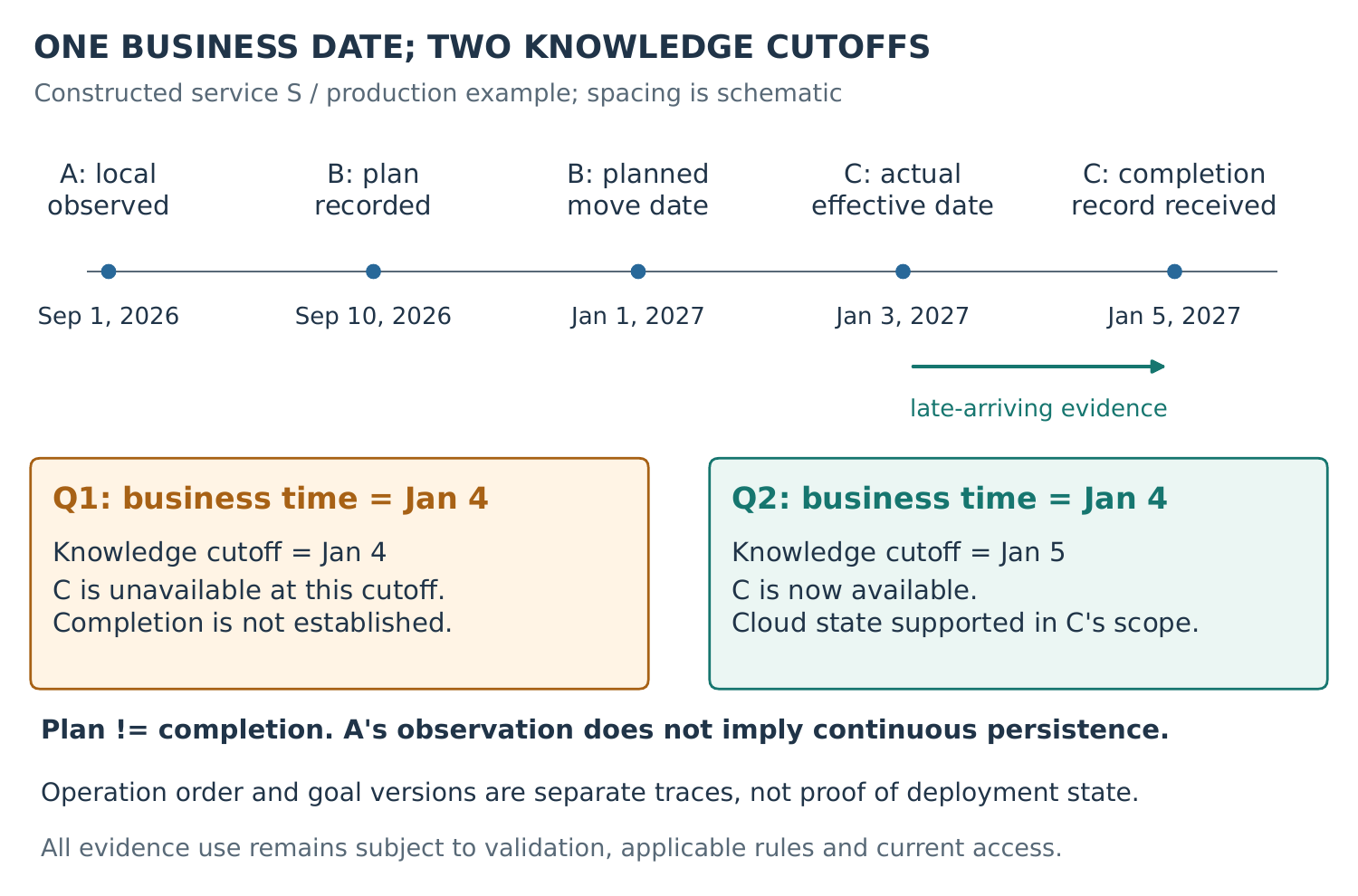}

\par\small\raggedright

\textbf{Figure 3. One business date, two knowledge cutoffs.} A
completion effective on January 3 becomes known on January 5. For the
January 4 business-state query, the January 4 cutoff cannot use that
evidence; the January 5 cutoff can. The plan does not prove completion,
and an access trace does not establish deployment state. Dates and
spacing are illustrative, not a measured timeline or an assumption of
continuous on-premises validity.

\end{minipage}\par\medskip

\subsection{6.2 Revocation Before Delayed
Execution}\label{revocation-before-delayed-execution}

Consider a synthetic export workflow in an authorized test environment.
A user may enqueue a job, an administrator may subsequently revoke
access, and a worker may execute the queued job later. Individually
plausible responses do not determine whether the complete sequence is
permitted. We stipulate a policy requiring authorization to be rechecked
at execution, with a declared revocation-propagation deadline. An
observed protected result after that deadline is a candidate violation
only if the evidence also establishes the same principal, resource,
relevant policy version, and absence of an applicable delegation
exception.

\Needspace{16\baselineskip}
\textbf{Table 4. A constructed temporal witness and its missing
premises.} This table is a validation design, not an exploit recipe or
an observed vulnerability.

{\begingroup\small
\begin{longtable}[]{@{}
  >{\raggedright\arraybackslash}p{(\linewidth - 6\tabcolsep) * \real{0.0700}}
  >{\raggedright\arraybackslash}p{(\linewidth - 6\tabcolsep) * \real{0.2300}}
  >{\raggedright\arraybackslash}p{(\linewidth - 6\tabcolsep) * \real{0.3500}}
  >{\raggedright\arraybackslash}p{(\linewidth - 6\tabcolsep) * \real{0.3500}}@{}}
\toprule\noalign{}
\begin{minipage}[b]{\linewidth}\raggedright
Step
\end{minipage} & \begin{minipage}[b]{\linewidth}\raggedright
Observation
\end{minipage} & \begin{minipage}[b]{\linewidth}\raggedright
Additional evidence needed
\end{minipage} & \begin{minipage}[b]{\linewidth}\raggedright
Index decision
\end{minipage} \\
\midrule\noalign{}
\endhead
\bottomrule\noalign{}
\endlastfoot
A & A job is accepted while access is valid. & Principal, resource, job
identity, and policy version. & Record acceptance; do not infer
completion or perpetual authority. \\
B & A revocation is recorded before execution. & Effective time,
propagation deadline, affected scope, and ordering evidence. & Revise
the applicable authorization interval; preserve the earlier valid
event. \\
C & A delayed execution yields a protected result. & Execution receipt,
result provenance, and applicable exceptions. & Link a candidate
witness; a response alone does not prove a policy violation. \\
D & Reset-and-replay reproduces the result. & Independent policy
adjudication and benign, reordered, and corrected controls. & Report a
confirmed test-case violation only if the oracle is satisfied. \\
\end{longtable}
\endgroup
}

If the approved policy instead grants durable authority at enqueue time,
the same sequence may be legitimate. If execution is within an allowed
propagation interval, the example does not yet establish a violation. If
ordering, identity, or the oracle is unknown, the result remains
inconclusive. This negative control prevents the index from
manufacturing vulnerabilities by treating every change of state as a
security failure.

The 5W1H organization exposes reusable questions: What response and
effect occurred? How do guards, actions, and effects relate? When did
authority change relative to execution? Who acted, and Where was the
trust boundary? Why was an action intended, if the source actually says
so? Which versioned policy interpretation applies? What-else searches
for omitted evidence or alternative interpretations; What-if explores a
reordered or differently authorized trace without changing the factual
record.

The validation loop is: retain authorized observations; propose
transition and context bindings; join only identity- and
order-compatible records; evaluate a separately approved policy; replay
and minimize a candidate witness; revise the state abstraction when
counterexamples refute it. A counterexample can invalidate the model
rather than expose a vulnerability. Known-CVE chains, cross-step
business-logic flaws, and ordering or concurrency defects must be
labeled separately. Open-source deployment or a vulnerable component
name alone establishes neither reachability nor exploitability.

\par\medskip\noindent\begin{minipage}{\linewidth}\centering

\includegraphics[width=\linewidth]{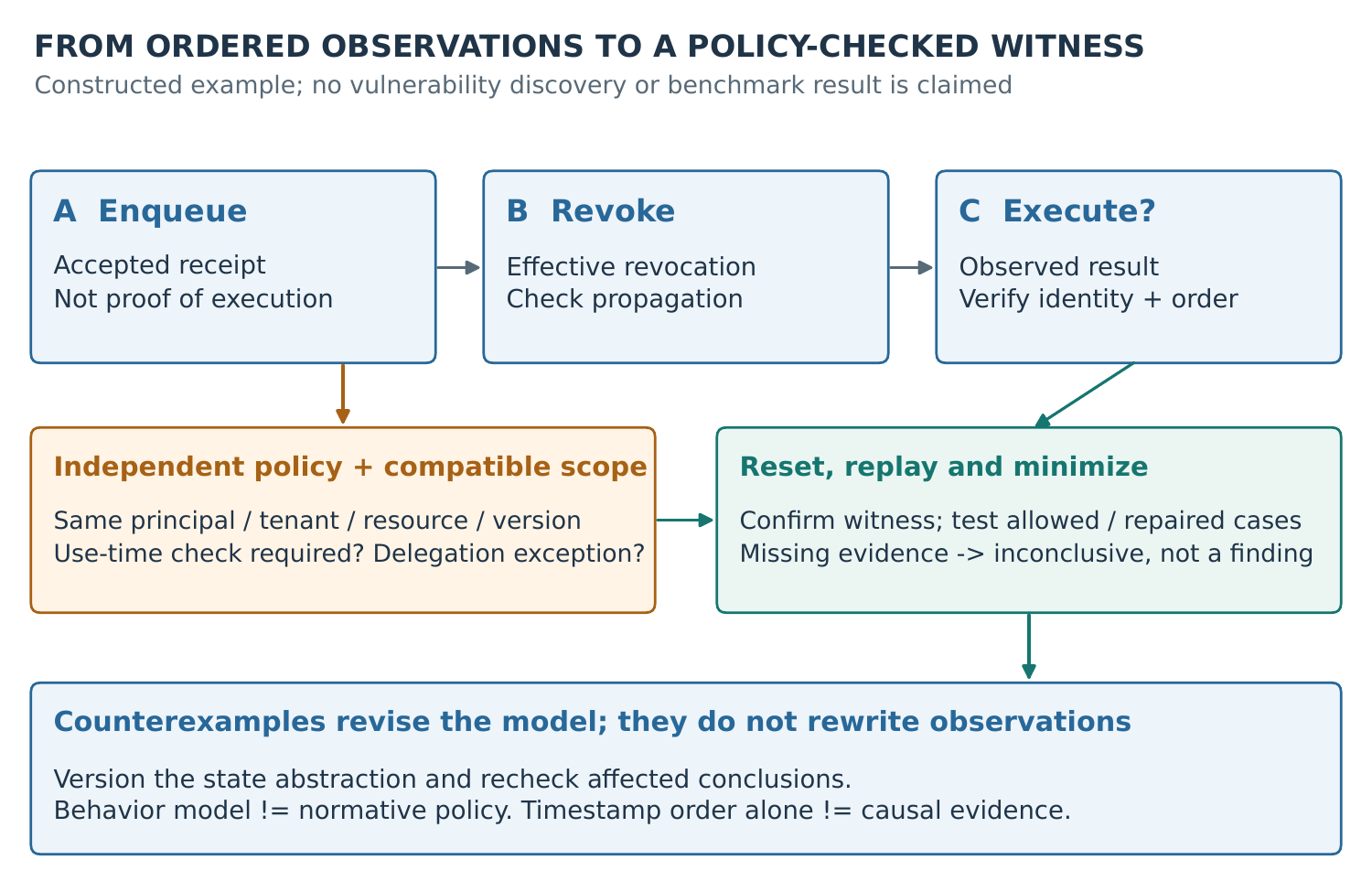}

\par\small\raggedright

\textbf{Figure 4. Evidence before a security conclusion.} A constructed
enqueue, revoke, and delayed-execution trace is joined with an
independently approved policy. Context and temporal checks produce a
candidate, not a finding. Reset-and-replay and negative controls
adjudicate it; counterexamples can revise the indexed state abstraction.
All timing is illustrative.

\end{minipage}\par\medskip

\section{7 Falsifiable Research
Hypotheses}\label{falsifiable-research-hypotheses}

\subsection{H1: Deferred Binding Improves Cross-Task Evidence
Coverage}\label{h1-deferred-binding-improves-cross-task-evidence-coverage}

With fixed sources, indexing budget, and initial tasks, allowing unbound
content to exist independently is expected to cover more valid evidence
for new tasks excluded from index design than retaining only information
accommodated by the current ontology. The measure is source-supported
information necessary for the task, not merely the number of nodes.

If the ontology is complete or a strong baseline also retains all
content, this difference may disappear. That would refute the stronger
claim that 5W1H inherently outperforms all representations and support a
narrower explanation: benefits arise from avoiding premature information
discard, not necessarily from the six labels themselves.

\subsection{H2: Contextualized Binding Reduces
Misuse}\label{h2-contextualized-binding-reduces-misuse}

At matched answer coverage and available evidence, explicitly managing
business valid time, knowledge time, and scope is expected to reduce
stale-label errors, cross-environment misuse, and the treatment of plans
as facts. A system that lowers its error rate merely through extensive
abstention does not support this hypothesis.

If a typed fact graph with equivalent information and constraints
achieves the same result, the paper should acknowledge that the core
benefit is achievable through general context management rather than
being exclusive to 5W1H.

\subsection{H3: Dependency Maintenance Reduces the Cost of Correct
Updates}\label{h3-dependency-maintenance-reduces-the-cost-of-correct-updates}

For local changes to content, bindings, or rules, dependency maintenance
is expected to reduce the scope of revalidation and recomputation,
conditional on updated results agreeing with full recomputation. Both
withdrawal of old conclusions and discovery of new conclusions must be
tested. Measuring only the speed of invalidation marking does not
establish effective complete maintenance.

Incremental maintenance may offer no benefit when a change affects the
entire ontology, dependencies are highly dense, or authorization
boundaries change broadly. Performance advantages should be reported
only for explicitly defined change types and scales.

\section{8 Experimental Design}\label{experimental-design}

\subsection{8.1 Data and Tasks}\label{data-and-tasks}

The first evaluation is planned around synthetic scenarios and licensed
public records in IT operations and manufacturing quality, covering
deployment changes, responsibility changes, equipment measurements, and
rule versions. Synthetic data enable controlled counterexamples, while
real records test representational complexity. Results will be reported
separately; synthetic outcomes cannot substitute for real-world business
validity.

Tasks include static content lookup, evidence retrieval for previously
unknown tasks, bitemporal state queries, cross-context disambiguation,
conclusion maintenance after changes, and hypothetical-branch isolation.
Sources will be grouped by event or entity history, with separation by
time period and task template. The index will be frozen before test
questions are revealed, preventing those questions from being
retroactively incorporated into extraction prompts.

\subsection{8.2 Strong Baselines and
Ablations}\label{strong-baselines-and-ablations}

Comparators must include rich summaries preserving source locations,
static ontology indexes, dynamic ontology indexes that can retain open
content, and typed fact graphs containing the same claims, evidence,
contexts, and versions. SocraticKG-style 5W1H extraction may serve as an
extraction control. Without replication using the original configuration
and data, however, it must be described as an adapted baseline, and the
original paper's results must not be reused as results of this project.

The design must separate prompting from storage mechanisms. One
comparison holds underlying storage and binding fixed while comparing
5W1H prompting against open-ended atomic-fact extraction. A second fixes
the same atomic facts and compares organization with and without 5W1H
groupings. A third fixes extraction and compares early filtering with
progressive binding. A fourth provides identical temporal information
and compares merely storing those fields with actually enforcing
contextual admission. These comparisons prevent additional content,
larger budgets, or stronger validators from being misattributed to the
representation itself.

Methods should share model versions, evidence-access permissions, rule
expressivity, and evaluation questions. Necessary engineering
differences must be recorded. Domain ontologies and human checks can be
standardized as controlled inputs, but their costs must also be
accounted for separately. The proposed method must not receive full
context while baselines are deprived of temporal or version fields.

\Needspace{16\baselineskip}
\textbf{Table 5. Controlled comparisons for attribution.} All rows
retain common authorization protections; unsafe variants run only in a
synthetic, non-executing test harness. Each row isolates a factor,
rather than comparing an unprotected minimal system with a fully
engineered one.

{\begingroup\small
\begin{longtable}[]{@{}
  >{\raggedright\arraybackslash}p{(\linewidth - 6\tabcolsep) * \real{0.1500}}
  >{\raggedright\arraybackslash}p{(\linewidth - 6\tabcolsep) * \real{0.2500}}
  >{\raggedright\arraybackslash}p{(\linewidth - 6\tabcolsep) * \real{0.2700}}
  >{\raggedright\arraybackslash}p{(\linewidth - 6\tabcolsep) * \real{0.3300}}@{}}
\toprule\noalign{}
\begin{minipage}[b]{\linewidth}\raggedright
Test
\end{minipage} & \begin{minipage}[b]{\linewidth}\raggedright
Hold fixed
\end{minipage} & \begin{minipage}[b]{\linewidth}\raggedright
Vary
\end{minipage} & \begin{minipage}[b]{\linewidth}\raggedright
Primary outcome
\end{minipage} \\
\midrule\noalign{}
\endhead
\bottomrule\noalign{}
\endlastfoot
Extraction & Sources, model, token and storage budgets & 5W1H prompt
versus open atomic-fact prompt & Evidence coverage and qualifier
retention \\
Binding & Same extracted content and mapper & Early schema filtering
versus retained unbound content & New-task evidence coverage at matched
precision \\
Categories & Same facts, lifecycle metadata, gates and routing & Named
categories versus a flat typed representation & Incremental calibration
and coverage; report equivalence margins if tested \\
Admission & Same facts, time/scope metadata and access controls & Stored
metadata versus enforced semantic gates & Contextual misuse at matched
answer coverage \\
Applicability & Same evidence and query templates & Positive
indirect-use cases versus scope-mismatched negative pairs & Useful
activation and false activation \\
Maintenance & Same event stream, semantics and measured cost accounting
& Full recomputation versus incremental repair & Withdraw/add/retain
agreement and total cost \\
\end{longtable}
\endgroup
}

In particular, the categories test fixes policy inputs and routing: it
is not a replication of the multifactor ablation in {[}R16{]}. A
factorial follow-up can measure category-by-policy interactions, but
must be labeled separately. Added information, a larger effective
budget, or looser privacy constraints must not be credited to the 5W1H
name.

\subsection{8.3 Metrics, Statistics, and Failure
Interpretation}\label{metrics-statistics-and-failure-interpretation}

Coverage metrics include recall of task-required evidence, evidence
precision, and retention of qualifiers such as values, units, negation,
and modality. Reliability metrics include binding correctness,
contextual misuse rate, answer coverage, and risk-coverage curves.
Update metrics compare all evaluated conclusions across old and new
snapshots, separately checking correct withdrawal, retention, and
addition. Costs include initial extraction, binding revalidation, human
annotation, querying, updating, and storage.

Following the distinctions in {[}R30, R31{]}, evaluate four layers
separately: stored state, task-level evidence views, agent use, and
final outcomes. After ingestion, inspect current and historical records;
at a fixed snapshot, assess required-evidence coverage and incorrect
joins; then test when memory should be ignored, bounded in scope, or
helpfully used. This separates stored-but-unused evidence, correctly
retrieved but misused evidence, and incorrect state that happens to
yield a correct answer. FEB30 is a small synthetic benchmark, not
evidence of enterprise-scale validity {[}R31{]}.

At least two independent annotators should be used, with adjudication of
disagreements and disclosure of the evidence available to evaluators.
Confidence intervals should resample source histories or event groups,
rather than treating many questions about the same event as independent
samples. Final sample sizes, primary metrics, thresholds, and
practically meaningful improvement criteria should be frozen after a
pilot and before the test set is revealed. Detailed procedures are
provided in the companion \emph{Experiment and Implementation Protocol}
(Chinese; see the repository README).

The proposed datasets have not yet been constructed and these
comparisons have not been run. Accordingly, this manuscript contains no
results tables, performance improvements, or significance claims. Even
if the full method outperforms weak baselines, strong baselines and
ablations must identify the source of the gain. If the benefit is solely
engineering integration, it should be reported as an engineering
contribution.

For revocation scenarios, report three denominators explicitly: the
fraction of trials exposing an invalidated record to the agent, the
fraction admitting it as a premise, and the fraction proposing an unsafe
simulated action. Authorized historical retrieval is not automatically a
failure; the requested mode and allowed uses determine the label.
Access-policy violations are evaluated separately from semantic misuse.
Track the end-to-end answered coverage and risk curve so that blocking
everything cannot count as success. Maintenance reporting includes time
spent unable to establish a current answer, not merely the latency of
completed updates.

\subsection{8.4 Optional Security
Substudy}\label{optional-security-substudy}

This extension is not required to validate the general index and must
not be counted as an already demonstrated contribution. \textbf{HS1}
tests whether context-valid transition joins reduce false accusations at
matched candidate coverage relative to field-equivalent trajectory
graphs. \textbf{HS2} tests whether counterexample-driven
state-abstraction revision improves confirmed held-out policy-violation
coverage over a fixed abstraction, after charging model-revision and
replay costs. A null result narrows the contribution to representation
and integration.

Use isolated, explicitly authorized workflows with independently labeled
normative policies. Include use-time reauthorization, durable
enqueue-time delegation, bounded propagation, late observations,
reordered events, and tenant or resource mismatches. Separate known-CVE
composition from non-CVE logic and temporal cases. Split by application
and workflow family, not merely by event or prompt; disclose
public-example contamination and report synthetic versus real cases
separately.

Compare against a typed trajectory graph with the same facts, policies,
LLM, tools, and budget; add fixed-state and context-ablated variants.
RESTler-like stateful dependency exploration and LogiAgent-like logical
testing are relevant adapters where interfaces permit them; ChatAFL is a
protocol-specific comparator, not a drop-in baseline for every
application. Report adaptation effort and unsupported cases rather than
asserting an unrun head-to-head comparison.

Count unique, independently adjudicated policy violations,
benign-workflow false positives, inconclusive outcomes, and minimum
reproducible witness length. Charge extraction, exploration, execution,
policy review, state revision, and replay to total cost; report latency
and safe-task coverage. More candidate alerts, more generated tests, or
more refusals alone is not success. Tests T23--T30 in the companion
protocol specify the required negative controls.

\section{9 Extension: An Indexed Corpus for Training and
Evaluation}\label{extension-an-indexed-corpus-for-training-and-evaluation}

\subsection{9.1 Separating the Data Foundation from Downstream
Systems}\label{separating-the-data-foundation-from-downstream-systems}

The long-term purpose of 5W1H+Which is to provide reusable, extracted
material for improving agents. Its immediate product is an indexed
corpus, not a training set, reward function, or four separate buckets of
records. Independently addressable content units retain evidence,
context, and binding states, while evidenced cross-record relations
connect goals, actions, observations, evaluations, and resource use.
Temporal proximity or semantic similarity alone does not establish that
two records describe the same task or outcome.

Benchmarks, positive and negative examples (shots), trajectories, and
costs are four organizing pillars of a downstream training and
evaluation system. They are not four labels that every source record
must receive. Task-level evidence views assemble corpus material for a
particular task and goal version, referencing source versions, content
units, relation evidence, applicable bindings, selection rules, and
missing information. These views are reproducible selections and joins,
not additional sources of truth. Their dependencies are many-to-many:
one observation may support several views, while one usable example may
require several sources. Unbound material remains discoverable rather
than being discarded because it does not fit the current training
objective.

\par\medskip\noindent\begin{minipage}{\linewidth}\centering

\includegraphics[width=\linewidth]{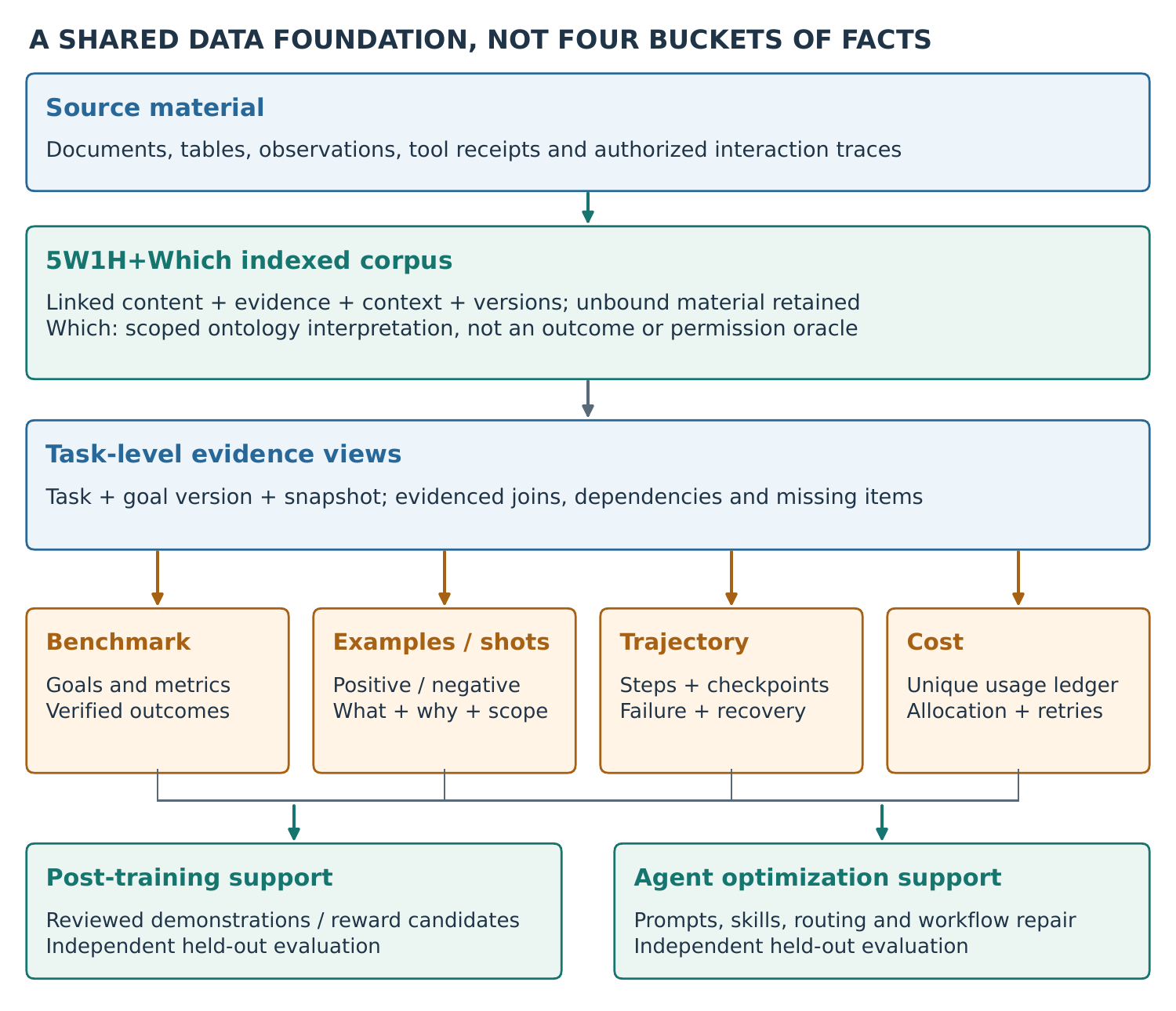}

\par\small\raggedright

\textbf{Figure 5. From a reusable indexed corpus to training and
evaluation.} The corpus retains content, relations, evidence, and
validity; task and goal versions govern material assembly. Four
connected pillars support both post-training and non-parametric agent
optimization. Arrows denote proposed information dependencies, not an
implemented training pipeline or measured capability gains.

\end{minipage}\par\medskip

\Needspace{20\baselineskip}
\subsection{9.2 Four Pillars, Not One-to-One Field
Assignments}\label{four-pillars-not-one-to-one-field-assignments}

\textbf{Table 6. Shared corpus support for four downstream pillars.} The
listed 5W1H directions are useful entry points, not exclusive field
assignments. Which checks the applicable ontological interpretation;
independent validation is still required for the decisions in the final
column.

{\begingroup\small
\begin{longtable}[]{@{}
  >{\raggedright\arraybackslash}p{(\linewidth - 6\tabcolsep) * \real{0.1500}}
  >{\raggedright\arraybackslash}p{(\linewidth - 6\tabcolsep) * \real{0.2500}}
  >{\raggedright\arraybackslash}p{(\linewidth - 6\tabcolsep) * \real{0.3200}}
  >{\raggedright\arraybackslash}p{(\linewidth - 6\tabcolsep) * \real{0.2800}}@{}}
\toprule\noalign{}
\begin{minipage}[b]{\linewidth}\raggedright
Pillar
\end{minipage} & \begin{minipage}[b]{\linewidth}\raggedright
Core question
\end{minipage} & \begin{minipage}[b]{\linewidth}\raggedright
Material assembled from the corpus
\end{minipage} & \begin{minipage}[b]{\linewidth}\raggedright
Independent decisions and checks
\end{minipage} \\
\midrule\noalign{}
\endhead
\bottomrule\noalign{}
\endlastfoot
Benchmark & What counts as completion, how is it compared, and why does
it matter to the business? & What goals, outputs, and metric
definitions; Why business relevance; context and validity scope & Metric
validity, thresholds, relative baseline, denominator, direction, and
priority \\
Examples & Which outputs are acceptable or unacceptable, under what
conditions, and why? & What inputs, outputs, and formats; Why stated
reasons or evidence; relevant context and process & Labels under a
frozen outcome oracle; format compliance alone is not correctness \\
Trajectory & How did the task progress, and how did local failures
affect completion? & Steps, checkpoints, dependencies, attempts,
recovery, goal revisions, and evidence & Identity, partial order,
checkpoint and final verdicts under frozen criteria; no invented steps
or causal links \\
Cost & Which resources were used to obtain the outcome? & When, Who, and
Where link resource use, actions, measurements, and price versions &
Measurement, units, pricing, allocation, retries, and failed-attempt
costs \\
\end{longtable}
\endgroup
}

A benchmark view distinguishes terminal business outcomes from process
indicators and proxies. Business priority requires an explicit
owner-approved rationale and goal version; correlation or an
LLM-generated ranking does not establish causal business value. Absolute
and relative metrics need stated units, windows, direction, and, where
relevant, a frozen comparator and budget. Extracted candidate metrics
remain candidates until independently approved.

ImpossibleRubrics motivates testing generated criteria against
independently decidable tasks and adversarial outputs rather than
treating a rubric as ground truth {[}R35{]}. Our proposed benchmark
protocol freezes the oracle version, distinguishes impossible,
undecidable, and insufficient-evidence cases, and scores output format,
stated rationale, and business success separately. The same LLM must not
be the sole label creator, optimizer, and final judge.

Positive and negative labels are conditional on the task, context,
required output, and evaluation oracle; they are not permanent
properties of a fact. Missing evidence yields an unknown label, not an
automatic negative. Why distinguishes user-stated reasons,
contemporaneous system explanations, and retrospective hypotheses. A
contrast between a successful and failed example may motivate analysis
but does not by itself identify a causal reason for the difference.

Costs require a common usage ledger with unique entries, measurement or
estimation status, units, and pricing versions. The same tool call may
appear in benchmark, example, and trajectory views without being charged
three times. Shared resources require an explicit allocation rule, and
unallocated overhead remains visible. Reporting should include failed
attempts, retries, and escalations, alongside completion rate and cost
per verified successful task; cheap failures must not be presented as
efficiency gains.

PeakBench distinguishes logical tool planning from physical resource
scheduling {[}R34{]}. We accordingly propose retaining dependencies and
measurement windows, while reporting tool calls, tokens, monetary cost,
elapsed latency, and peak resource use separately. Parallel-call
durations cannot simply be summed as end-to-end latency, and
heterogeneous resource measures cannot be added without a justified
conversion. These quantities require external measurements; Which cannot
certify them by semantic matching.

\subsection{9.3 Process Evidence, Goal Revision, and View
Maintenance}\label{process-evidence-goal-revision-and-view-maintenance}

Business facts are not the only source of time. We distinguish fact
valid time, system knowledge time, the order or partial order of
operational events, and goal version. A goal version is not another
calendar clock, and concurrent operations cannot necessarily be ordered
by a single timestamp. An episode can record visible context, the
current goal, tools and arguments, returned results, deviation from the
plan, user revisions, and subsequent actions. What identifies the
intended or observed state; How preserves the process leading to it.
Observed traces, reconstructed traces, and simulated alternatives must
remain distinguishable.

The useful object is a linked evidence structure: goals and metrics
connect to cases, cases to trajectories, and trajectories to usage
records, with evidence, scope, and version for each relation. Which can
validate an ontological interpretation, but it does not prove event
identity, factual truth, authorization, task success, or a measured
cost. These checks remain explicit external responsibilities.

The dependency mechanism in Section 5 can be extended to task-level
views. Source corrections, goal revisions, binding changes, and revised
outcome criteria may invalidate selections or labels and introduce new
eligible material. A view update must therefore support additions as
well as withdrawals. Historical labels remain tied to their original
goal and oracle versions; unresolved current cases become pending or
unknown. Evaluation and training exports need a consistent snapshot
rather than an accidental mixture of versions.

\subsubsection{9.3.1 Trajectory as a Versioned Process-Evidence
Graph}\label{trajectory-as-a-versioned-process-evidence-graph}

A trajectory is an episode-specific process record, not a skill and not
merely an ordered tool log. Its proposed representation connects step
attempts, checkpoint assessments, evidence dependencies, recovery
relations, and task-level verdicts under a task identity, goal version,
and observation cutoff. A step retains its acting principal,
decision-time context, intended action, arguments, receipt, observed
effect, and evidence references. Attempt identity prevents retries from
overwriting earlier failures. Distributed steps retain supported partial
orders; unsupported ordering and missing observations remain explicit.

A checkpoint expresses a task requirement or an authorized intermediate
acceptance condition, not necessarily one tool call or one step. Several
steps may support one checkpoint; one step may affect several
checkpoints. A frozen checkpoint specification records its criterion,
scope, verification route, applicability, criticality, and dependencies.
Its assessment separately records satisfied, violated, or unknown, with
evidence and evaluator version. Not-applicable requires an explicit
applicability decision; it is not a substitute for missing evidence.
Criteria derived retrospectively must be marked as retrospective and
cannot be presented as prospectively frozen evaluation.

\Needspace{16\baselineskip}
\textbf{Table 7. Three distinct decision layers in a trajectory.} Each
layer retains its own evidence and version; success at one layer does
not imply success at another.

{\begingroup\small
\begin{longtable}[]{@{}
  >{\raggedright\arraybackslash}p{(\linewidth - 4\tabcolsep) * \real{0.1800}}
  >{\raggedright\arraybackslash}p{(\linewidth - 4\tabcolsep) * \real{0.4000}}
  >{\raggedright\arraybackslash}p{(\linewidth - 4\tabcolsep) * \real{0.4200}}@{}}
\toprule\noalign{}
\begin{minipage}[b]{\linewidth}\raggedright
Layer
\end{minipage} & \begin{minipage}[b]{\linewidth}\raggedright
Record and evidence
\end{minipage} & \begin{minipage}[b]{\linewidth}\raggedright
Decision boundary
\end{minipage} \\
\midrule\noalign{}
\endhead
\bottomrule\noalign{}
\endlastfoot
Step attempt & Context, action, receipt, effect, actor, time, and
attempt identifier & Tool execution status does not certify a business
requirement; absent effects remain unknown \\
Checkpoint & Frozen requirement, verification route, supporting steps,
applicability, and assessment & Satisfied, violated, or unknown under
the stated criterion; critical gates are not averaged away \\
Task outcome & Goal and oracle versions, required checkpoints,
constraints, and independent final evidence & Pass, fail, or unknown
according to the authorized acceptance rule; process quality is reported
separately \\
\end{longtable}
\endgroup
}

GCPC provides direct prior art for evidence-grounded partial-credit
evaluation and separation from official outcomes {[}R36{]}. Long-Horizon
Agent Trajectory Attribution studies attribution across interacting
components {[}R37{]}; a chronological or dependency edge in our index is
not by itself a causal finding. TQS studies trajectory quality beyond
binary success for data selection {[}R38{]}, while Write Once, Check
Many studies reusable records for later policy checks {[}R39{]}.
Accordingly, neither process evaluation nor reusable trajectory
summaries is claimed as new here. The research question is whether
context-valid indexing improves their evidence assembly, reuse, and
versioned maintenance relative to field-equivalent baselines.

\subsubsection{9.3.2 Local Failure, Recovery, and Overall
Impact}\label{local-failure-recovery-and-overall-impact}

Local failure has conditional, not universal, consequences. A failed
attempt may be recovered by a later verified retry; an unresolved
required checkpoint may block completion; an optional reporting failure
may reduce quality without invalidating the business outcome. A
non-compensable safety or authorization violation may fail the task
despite a technically correct output. The acceptance policy must specify
which conditions are recoverable, required, optional, or non-compensable
before scoring. These roles cannot be inferred solely from the agent's
explanation.

We therefore retain an impact record linking a failed attempt or
checkpoint to affected requirements, blocked downstream steps, recovery
or compensation evidence, final verdict, and the rule authorizing that
interpretation. Recovery creates a new assessment; it does not erase the
original failure, harm, or cost. Compensation is not assumed to undo an
irreversible effect. Dependency reachability identifies potentially
affected records, not proof that an error caused all downstream
failures. Causal claims require additional interventions, controlled
replays, or independently reviewed evidence.

For diagnosis, report satisfied requirements among decidable applicable
checkpoints together with decidability coverage, unknown counts,
critical violations, recovery status, and the independent task verdict.
A high partial-credit fraction with low coverage must not look like
successful completion. Weighted variants require predeclared weights;
neither an average nor a recovered local error can override a
non-compensable gate. Repeated attempts at the same goal remain separate
episodes linked by task-family and goal versions, supporting comparisons
without synthesizing a fictional successful trace from different runs.

\par\medskip\noindent\begin{minipage}{\linewidth}\centering

\includegraphics[width=\linewidth]{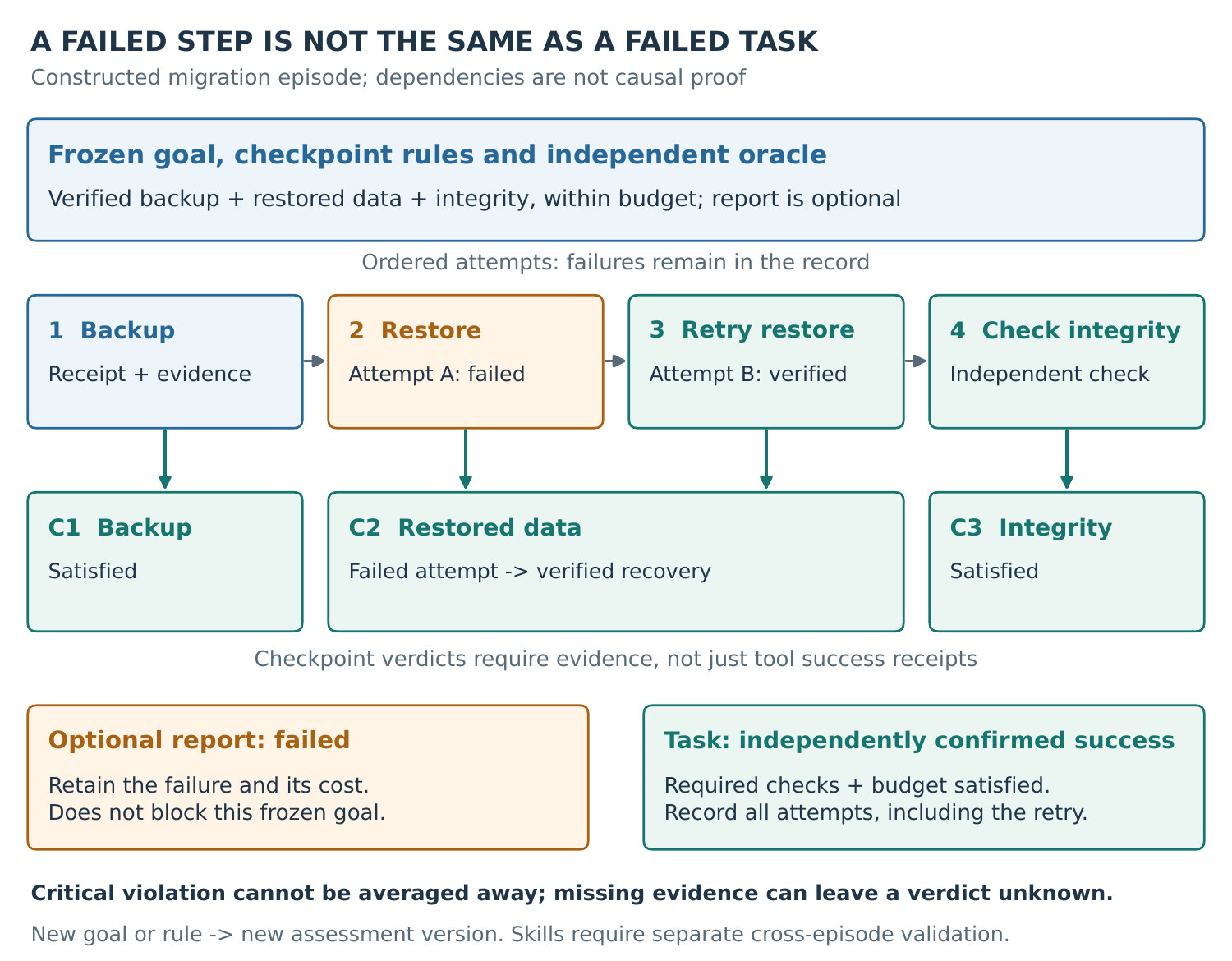}

\par\small\raggedright

\textbf{Figure 6. From attempts to checkpoint evidence and task-level
judgment.} In this constructed migration example, a verified retry
recovers a failed restore attempt, while an optional reporting failure
remains visible. Independent final checks decide the outcome under a
frozen goal and oracle. Earlier failures and their costs remain
recorded. A critical unresolved or non-compensable violation cannot be
offset by other passed checkpoints. Arrows show proposed evidence or
dependency links, not demonstrated causality or measured performance.

\end{minipage}\par\medskip

\subsection{9.4 Constructed Example: A Migration
Task}\label{constructed-example-a-migration-task}

Consider a synthetic task to migrate a service while satisfying
specified acceptance criteria and a resource budget. A migration plan
states the goal but does not establish completion. A tool receipt
records an action but not necessarily its business outcome. Acceptance
specifications, independent verification, and usage records supply
different parts of the evidence. No single source necessarily suffices
for all four pillars.

A benchmark view assembles acceptance metrics and verified outcomes.
Example views link input context and required formats to outputs labeled
under the same criteria. A trajectory view preserves the attempted
steps, initial failure, and subsequent verified retry. A cost view joins
unique resource entries to those attempts. If acceptance evidence is
absent, the case is not promoted to a positive example merely because a
tool returned success. If the user changes the goal, the system creates
a new versioned view rather than rewriting the historical task as if the
new requirement had always applied. This illustrates a proposed
dependency structure, not an empirical result.

More specifically, suppose the frozen acceptance rule requires a
verified backup, successful restore, and an independent integrity check
within budget. A failed first restore is followed by a verified retry.
The restore checkpoint can then be satisfied, while the first attempt
remains failed and both costs count. Failure to generate an optional
narrative report does not alone fail this task; if that report is
mandatory under another goal version, its failure does block completion
there. Missing integrity evidence yields an unknown final verdict unless
another independently verified failure already determines failure. A
confirmed non-compensable authorization breach cannot be repaired merely
by passing integrity checks. These contrasting cases test whether the
index preserves the acceptance rule and its evidential scope rather than
learning that any retry implies success.

\subsection{9.5 Supported Optimization and Evaluation
Boundaries}\label{supported-optimization-and-evaluation-boundaries}

Authorized and reviewed views may support demonstrations, contrastive
examples, student-error correction, candidate reward components, and
independently held-out evaluations. The learning algorithm and training
policy remain downstream choices. Extracted QA pairs do not
automatically constitute valid rewards, and self-generated answers must
not become their own gold standard. Mutable facts, permissions, prices,
and ontology versions remain externally checked rather than being
treated as permanently correct model knowledge.

The external-memory transfer studied in Agent Memory Distillation and
parameter learning studied in Self-Retrospection Distillation require
separate comparisons, with training and inference budgets disclosed
{[}R32, R33{]}. Contemporaneously available context must be
distinguished from later hindsight: retrospective material may inform
authorized training supervision, but cannot silently enter historical
decision inputs or reveal final-test answers.

Corpus retention and consumer selection also require separate policies.
JitMem's successful-trace memory does not imply that failed attempts are
useless for other consumers {[}R28{]}. Failures may support diagnosis,
negative examples, or retrospective supervision {[}R33{]} without being
suitable forward demonstrations. Views should specify allowed uses and
label provenance; success-only, failure-only, and mixed-material
ablations should test utility rather than assume that retaining more
material always helps.

The same corpus can support non-parametric agent optimization: prompt
revision, tool selection, failure analysis, skill candidates, workflow
repair, and escalation or stopping rules. InMind motivates retrieving
evidence beyond the current question's surface terms {[}R14{]}, while
Recuris already studies experience and working-memory updates {[}R15{]}.
We therefore do not claim to originate experience-to-skill learning. A
candidate skill must carry prerequisites, exclusions, provenance,
failure branches, and rollback conditions. Promotion requires validation
across independent held-out episodes, including counterexamples; one
successful trace neither establishes generalization nor authorizes
deployment.

Trajectory is the process-evidence pillar; a skill is a possible
downstream abstraction across reviewed trajectories. A failed trajectory
may still yield useful diagnostic material without becoming a successful
demonstration. Neither a checkpoint graph nor an automatically generated
skill is validated simply by being indexed.

Evaluation should first test the data foundation: completeness and
correctness of task-level evidence, reuse across the four pillars, and
view maintenance relative to a full rebuild. A strong baseline must
include a field-equivalent typed graph or relational representation with
joins, provenance, and versioned views, not merely isolated text
snippets. The companion protocol specifies proposed D1--D3 tests and
T31--T50 contracts, including checkpoints, failure impact, and recovery;
these extend rather than replace H1--H3, and report no training gains.

Source lineage, entity histories, and workflow families should be split
before generating views or examples. Materials reserved for final
evaluation must not feed training, reward design, or index optimization.
A shared corpus provides shared representation, not permission to expose
every test record to every consumer. Only after this data-level
evaluation should a separately controlled study test downstream agent
performance and its total cost.

\section{10 Limitations, Risks, and
Ethics}\label{limitations-risks-and-ethics}

5W1H does not provide inherently orthogonal dimensions. Who may describe
both a responsible party and a tool operator; Where may refer to a
physical location or a computational environment. Reducing the
representation to six text boxes may obscure role distinctions. Typed
payloads and domain extensions remain essential, and no finite index can
guarantee support for arbitrary future questions.

LLM extraction, identity disambiguation, and ontology mapping can all
fail. Evidence links improve auditability but do not prove the evidence
itself true. Validators may miss conflicts or be affected by malicious
instructions embedded in sources. Source content should be treated as
data to parse, not as instructions controlling system behavior or
expanding permissions. Consequential decisions require verification and
human review proportional to their risk.

Separating content, bindings, and dependencies increases storage,
version-management, and implementation complexity. If strong baselines
reach the same outcomes with the same budget, the proposal may be a
maintainable implementation convention rather than a novel theory of
representation. Richer indexes may also increase irrelevant retrieval,
so filtering cost and user burden must be measured.

User behavior logs may contain identities, intentions, and sensitive
business content. Collection and processing should follow explicit
authorization, purpose limitation, and data minimization. Only events
and fields necessary for the research should be retained, with access
isolation and deletion propagation. Unauthorized private records must
not be reused merely to ``generate a Skill.'' Public examples should be
synthetic or appropriately de-identified; de-identification must not be
equated with the absence of re-identification risk.

Reuse for training also requires purpose-specific authorization,
minimization, access isolation, and deletion propagation through
dependent views and exports. Removing an index entry does not unlearn
information already absorbed into model weights. Training lineage and
model versions therefore require a separate process for restricting use,
retraining, or other justified remediation. Extracting a metric,
labeling examples with it, and evaluating the same system against those
labels risks circular validation; an independent assessment boundary
must be retained.

\section{11 Conclusion}\label{conclusion}

We propose using 5W1H to organize evidenced content, Which to manage
revisable ontological interpretations, and contextual validity to govern
queries and updates. The central problem is not to find a fixed schema
replacing every industry ontology. It is to preserve information that is
not yet bound, make bound information usable under explicit conditions,
and trace the consequences when those conditions change.

The longer-term product is a reusable indexed corpus supporting
benchmarks, examples, trajectories, and costs as connected pillars of
training and evaluation, not four collections of isolated facts. Both
post-training and non-parametric agent optimization can consume
task-level evidence views while retaining separate validation and
authorization responsibilities.

This is a method and experimental-design manuscript. The immediate
priorities remain a fair comparison with a fact graph containing the
same information and tests of the three hypotheses on cross-task
evidence coverage, contextual misuse, and update cost, followed by tests
of task-level view reuse and maintenance. Only after obtaining that
evidence can we assess whether the mechanism provides a sufficient
contribution for an independent paper.

\section{References}\label{references}

Selection emphasizes direct prior art and primary sources. The October
7, 2026 follow-up adds {[}R19{]}--{[}R26{]}; the October 22 follow-up
adds nine fixed-version sources {[}R27{]}--{[}R35{]} on ontology
evolution, memory, learning, and evaluation. The October 23 refinement
adds four trajectory sources {[}R36{]}--{[}R39{]}. Venue history,
artifact recognition, and repository attention inform reading priority,
not truth or novelty. No uniform verified citation-count dataset or
time-series evidence of trending growth was available; none is asserted
here. Companion notes record dated attention snapshots and reading
depth.

\begin{itemize}
\setlength{\itemsep}{3pt}\setlength{\parsep}{0pt}
\item
  {[}R1{]} Sanghyeok Choi, Woosang Jeon, Kyuseok Yang, and Taehyeong
  Kim. \emph{SocraticKG: Knowledge Graph Construction via QA-Driven Fact
  Extraction}. Findings of ACL 2026, pp.~39149-39169.
  \href{https://aclanthology.org/2026.findings-acl.1951/}{ACL
  Anthology}; \href{https://arxiv.org/abs/2601.10003}{arXiv:2601.10003}.
  The preprint first appeared in January 2026; conference publication in
  July must not be treated as the method's first appearance.
\item
  {[}R2{]} Rajesh Piryani, Nathalie Aussenac-Gilles, Nathalie Hernandez,
  Cédric Lopez, and Camille Pradel. \emph{Ontology Based Event Knowledge
  Graph Enrichment Using Case Based Reasoning}. SEMANTiCS 2024 /
  \emph{Knowledge Graphs in the Age of Language Models and
  Neuro-Symbolic AI}.
  \href{https://journals.sagepub.com/doi/10.3233/SSW240017}{Publisher
  page}; \href{https://github.com/rpiryani/xpEventCore}{authors'
  xpEventCore project}.
\item
  {[}R3{]} Xiaohui Zhang, Zequn Sun, Chengyuan Yang, Yuanning Cui,
  Lingbing Guo, and Wei Hu. \emph{Toward Effective and Reliable LLM
  Agents via Dynamic Ontology}. arXiv:2608.22974, August 24, 2026.
  \href{https://arxiv.org/abs/2608.22974}{Preprint}.
\item
  {[}R4{]} Siddhesh Thombre, Manasi Patwardhan, and Sunita Sarawagi.
  \emph{Surprising Effectiveness of Self-Demonstrations in Enhancing
  Schema-Ontology Mapping with LLMs}. arXiv:2609.13776, September 12,
  2026. \href{https://arxiv.org/abs/2609.13776}{Preprint}.
\item
  {[}R5{]} Zhangcheng Qiang, Kerry Taylor, and Weiqing Wang.
  \emph{OM4OV: Leveraging Ontology Matching for Ontology Versioning}.
  Transactions on Graph Data and Knowledge 4(2), 6:1-6:19, September 3,
  2026.
  \href{https://drops.dagstuhl.de/entities/document/10.4230/TGDK.4.2.6}{Publisher
  page}. This date denotes journal publication, not necessarily the
  method's first appearance.
\item
  {[}R6{]} FBK Data and Knowledge Management. \emph{CKR - Contextualized
  Knowledge Repository}.
  \href{https://dkm.fbk.eu/technologies/theoretical-frameworks/ckr-contextualized-knowledge-repository/}{Project
  technical description}. An institutional technical page, not a recent
  conference paper verified as such for this manuscript.
\item
  {[}R7{]} W3C. \emph{PROV-O: The PROV Ontology}.
  \href{https://www.w3.org/TR/prov-o/}{Standard}.
\item
  {[}R8{]} W3C. \emph{Time Ontology in OWL}.
  \href{https://www.w3.org/TR/owl-time/}{Standard}.
\item
  {[}R9{]} W3C. \emph{OWL 2 Web Ontology Language Primer (Second
  Edition)}. \href{https://www.w3.org/TR/owl2-primer/}{Standard}.
\item
  {[}R10{]} Furqan Nasir, Muhammad Atif Saeed, Muhammad Ehsan, Sher Jeel
  Ahmad, and Abdul Moiz Altaf. \emph{OntoKG-EQ: A provenance-grounded,
  competency-question-governed knowledge graph for auditable analyst
  querying}. arXiv:2609.08869, September 8, 2026.
  \href{https://arxiv.org/abs/2609.08869}{Preprint}.
\item
  {[}R11{]} Yucheng Wang et al.~\emph{The Illusion of What If:
  Evaluating the Breakdown of Counterfactual Reasoning in LLMs}.
  arXiv:2608.27953, August 28, 2026.
  \href{https://arxiv.org/abs/2608.27953}{Preprint}.
\item
  {[}R12{]} Wentao Qiu et al.~\emph{DimMem: Dimensional Structuring for
  Efficient Long-Term Agent Memory}. arXiv:2605.15759.
  \href{https://arxiv.org/abs/2605.15759v3}{Version 3}. A relevant May
  predecessor, not a September release.
\item
  {[}R13{]} Xingyuan Zeng et al.~\emph{RuleMem: Active Rule Memory for
  Long-Term Conversational Agents}. arXiv:2609.03915.
  \href{https://arxiv.org/abs/2609.03915}{Preprint}.
\item
  {[}R14{]} Ruizhe Li et al.~\emph{Keep It InMind: Benchmarking the
  Implicit-Association Blind Spot in Agent Memory}. arXiv:2607.24368.
  \href{https://arxiv.org/abs/2607.24368}{Preprint}.
\item
  {[}R15{]} Zhaochen Yu et al.~\emph{Recursive Experiential-Working
  Memory Evolution for Long-Horizon Agent Harnesses}. arXiv:2608.24876.
  \href{https://arxiv.org/abs/2608.24876}{Preprint}.
\item
  {[}R16{]} Ansuman Mullick and Eray Tüzün. \emph{Fortunate Recall:
  Ontology-Driven Memory Lifecycle Management for Persistent Coherence
  in LLMs}. arXiv:2609.10413, September 9, 2026.
  \href{https://arxiv.org/abs/2609.10413}{Preprint}.
\item
  {[}R17{]} Yi Ting Shen, Kentaroh Toyoda, and Alex Leung. \emph{Revoked
  but Still Authoritative: An Empirical Study of Revocation Enforcement
  in Agent-Memory Systems}. arXiv:2609.08258, September 8, 2026.
  \href{https://arxiv.org/abs/2609.08258}{Preprint}.
\item
  {[}R18{]} Beining Wu, Zihao Ding, and Jun Huang. \emph{ERRAND:
  Budgeted Maintenance of Agent Memory}. arXiv:2609.29545.
  \href{https://arxiv.org/abs/2609.29545}{Preprint}.
\item
  {[}R19{]} Ai et al.~\emph{A large language model-driven semantic
  inference framework for business logic vulnerability discovery in
  power system microservice APIs}. Journal of Engineering and Applied
  Science 73, 243, June 26, 2026.
  \href{https://link.springer.com/article/10.1186/s44147-026-01100-9}{Publisher}.
\item
  {[}R20{]} Ke Zhang et al.~\emph{LogiAgent: Automated Logical Testing
  for REST Systems with LLM-Based Multi-Agents}. arXiv:2503.15079, March
  19, 2025. \href{https://arxiv.org/abs/2503.15079}{Preprint}.
\item
  {[}R21{]} Jiangrong Wu, Huaijin Wang, Yihao Zhang, Yuhong Nan, and
  Shuai Wang. \emph{MOSAIC: Knowledge-Guided CLI Command Composition
  Attack in LLM Coding Agents}. arXiv:2607.02857, July 3, 2026.
  \href{https://arxiv.org/abs/2607.02857}{Preprint}.
\item
  {[}R22{]} Ashwini Kurady, Sri Sai Charith Grandhi, Rajesh Gupta, and
  Sumit Kumar. \emph{Compositional Policy Violations: When Step-Level
  Compliance Fails In Agentic AI Workflows}. arXiv:2609.18820, September
  16, 2026. \href{https://arxiv.org/abs/2609.18820}{Preprint}.
\item
  {[}R23{]} Vaggelis Atlidakis, Patrice Godefroid, and Marina
  Polishchuk. \emph{RESTler: Stateful REST API Fuzzing}. ICSE, 2019.
  \href{https://github.com/microsoft/restler-fuzzer}{Official
  implementation and paper links}.
\item
  {[}R24{]} Ruijie Meng, Martin Mirchev, Marcel Boehme, and Abhik
  Roychoudhury. \emph{Large Language Model Guided Protocol Fuzzing}.
  NDSS, 2024.
  \href{https://www.ndss-symposium.org/ndss-paper/large-language-model-guided-protocol-fuzzing/}{Conference}.
\item
  {[}R25{]} Gelei Deng et al.~\emph{PentestGPT: Evaluating and
  Harnessing Large Language Models for Automated Penetration Testing}.
  USENIX Security, 2024.
  \href{https://www.usenix.org/conference/usenixsecurity24/presentation/deng}{Conference}.
\item
  {[}R26{]} Giancarlo Pellegrino and Davide Balzarotti. \emph{Toward
  Black-Box Detection of Logic Flaws in Web Applications}. NDSS, 2014.
  \href{https://www.eurecom.fr/en/publication/4207}{Author institution}.
\item
  {[}R27{]} Meiduo Chong, Shaolei Zhang, Ju Fan, and Xiaoyong Du.
  \emph{EvoOntology: A Self-Evolving Ontology Layer for Data Agents}.
  arXiv:2609.15779v1, September 14, 2026.
  \href{https://arxiv.org/abs/2609.15779v1}{Fixed-version preprint}.
\item
  {[}R28{]} Yefan Zhou, Yang Li, Zeyu Leo Liu, Semih Yavuz, and Shafiq
  Joty. \emph{Just-in-Time Memory: Learning to Curate Task-Adaptive
  Memory for LLM Agents}. arXiv:2609.27334v1, September 23, 2026.
  \href{https://arxiv.org/abs/2609.27334v1}{Fixed-version preprint}.
\item
  {[}R29{]} Bowen Qin and Yao Lu. \emph{MoM: Memory of Memory}.
  arXiv:2609.25054v1, September 7, 2026.
  \href{https://arxiv.org/abs/2609.25054v1}{Fixed-version preprint}.
\item
  {[}R30{]} Ruike Cao, Fanyu Zhao, Fugen Yao, Liang Dong, Jian Xu, Yifei
  Zhao, Han Zhang, and Li Xiao. \emph{MemCalib: Benchmarking and
  Optimizing Memory Use in LLM Agents}. arXiv:2609.24259v3. First
  submitted September 21, 2026; cited revision October 8, 2026.
  \href{https://arxiv.org/abs/2609.24259v3}{Fixed-version preprint}.
\item
  {[}R31{]} Sebastian-Antonio Toma, Jean de Kernier, Mike Ply, and
  Oliviu Matei. \emph{Beyond Downstream QA: Rethinking How We Evaluate
  Agent Memory}. arXiv:2610.22209v1, October 15, 2026.
  \href{https://arxiv.org/abs/2610.22209v1}{Fixed-version preprint}.
\item
  {[}R32{]} Taeil Kim, Kangsan Kim, and Sung Ju Hwang. \emph{Agent
  Memory Distillation: Empowering Small LLM Agents with Hierarchical
  Teacher Memory}. arXiv:2608.07169v1, August 7, 2026.
  \href{https://arxiv.org/abs/2608.07169v1}{Fixed-version preprint}.
\item
  {[}R33{]} Haoxiang Zhang et al.~\emph{Self-Retrospection Distillation:
  Turning Post-hoc Experiences into Prior Foresight}.
  arXiv:2610.08077v3. First submitted October 6, 2026; cited revision
  October 14, 2026.
  \href{https://arxiv.org/abs/2610.08077v3}{Fixed-version preprint}.
\item
  {[}R34{]} Zhi-Kai Chen, Xu-Xiang Zhong, Song-Yan Li, De-Chuan Zhan,
  and Han-Jia Ye. \emph{PeakBench: Benchmarking Resource-Aware Tool
  Invocation in LLM Agents}. arXiv:2608.24509v1, August 25, 2026.
  \href{https://arxiv.org/abs/2608.24509v1}{Fixed-version preprint}.
\item
  {[}R35{]} Bowen Qin, Yi Xie, Yesheng Liu, and Xi Yang.
  \emph{ImpossibleRubrics: Stress-Testing Generated Rubrics as Reward
  Signals}. arXiv:2609.16816v1, September 15, 2026.
  \href{https://arxiv.org/abs/2609.16816v1}{Fixed-version preprint}.
\item
  {[}R36{]} Suliu Qin, Lu Yin, and Xilu Wang. \emph{Grounded Checklist
  Partial Credit for Agent Skill Trajectories}. arXiv:2608.27487v1,
  August 26, 2026.
  \href{https://arxiv.org/abs/2608.27487v1}{Fixed-version preprint}.
\item
  {[}R37{]} Jing Chen, Yang Sun, Li Zhang, Lin Xu, and Jie Shi.
  \emph{Long-Horizon Agent Trajectory Attribution: A Unified Benchmark
  and Fine-Grained Annotation Framework}. arXiv:2608.06909v1, August 7,
  2026. \href{https://arxiv.org/abs/2608.06909v1}{Fixed-version
  preprint}.
\item
  {[}R38{]} Sida Li, Yuanxin Liu, Shicheng Li, Linli Yao, Xu Sun, Haoli
  Bai, and Xiaohui Li. \emph{Quality, Not Just Outcome: A Structural
  Scoring Framework for Selecting Coding-Agent Trajectories for
  Supervised Fine-Tuning}. arXiv:2610.13152v1, October 9, 2026.
  \href{https://arxiv.org/abs/2610.13152v1}{Fixed-version preprint}.
\item
  {[}R39{]} Ashwin Nagarajan, Gursimran Panesar, Alicia Yang, Mona T.
  Diab, Virginia Smith, and Aashiq Muhamed. \emph{Write Once, Check
  Many: Reusable Ledgers for Post-Hoc Policy Compliance Auditing of
  Agent Trajectories}. arXiv:2610.24483v1, October 17, 2026.
  \href{https://arxiv.org/abs/2610.24483v1}{Fixed-version preprint}.
\end{itemize}

The October 23 trajectory refinement adds {[}R36{]}--{[}R39{]}, checked
against fixed-version primary sources for metadata and the claims used
here. It builds on the local trajectory reading notes; it is not an
exhaustive review, a popularity ranking, or an independent replication.

The earlier follow-ups retain their original scope: September 26, 2026
for {[}R1{]}--{[}R18{]} and October 7 for the security sources
{[}R19{]}--{[}R26{]}. The October 22 update targets
{[}R27{]}--{[}R35{]}; it does not imply that every older reference was
reverified. These are not systematic reviews or verified popularity
rankings. Reading depth, source versions, and revision decisions are
recorded in the companion literature notes and research log (Chinese;
linked from the README). The new preprints are related evidence, not
independently replicated findings or experimental results of this
project.
\end{document}